\pdfoutput=1
\RequirePackage{latexml}
\PassOptionsToPackage{table}{xcolor}
\PassOptionsToPackage{pagebackref,breaklinks}{hyperref}
\iflatexml
  \documentclass{article}
  \def\PAHTML{}
  \usepackage{amsmath,amssymb,bm,subcaption,xcolor}
  \usepackage[numbers,sort&compress]{natbib}
  \usepackage{hyperref}
  \usepackage[noabbrev,nameinlink]{cleveref}
  \usepackage{authblk}
  \let\affiliation\affil
\else
\documentclass{styles/bytedance_seed}
\fi

\usepackage{multirow}
\usepackage{array}
\usepackage{amsfonts}
\usepackage{nicefrac}
\usepackage{microtype}
\usepackage{xspace}
\ifdefined\PAHTML
  \usepackage{fancyvrb}
  \makeatletter
  \define@key{FV}{breaklines}{}
  \define@key{FV}{breakanywhere}{}
  \makeatother
\else
  \usepackage{fvextra}
\fi
\usepackage{placeins}
\ifdefined\PAHTML\else
  \usepackage{cuted} %
\fi

\providecommand{\PAbenchnote}{}
\providecommand{\PAintrofigs}{}
\providecommand{\PAdataexamples}{}
\providecommand{\PAqualitativeEarly}{}

\newcommand{\ours}{\textit{Paint-Anything}\xspace}
\DeclareRobustCommand{\hexcolor}[2][black]{\begingroup\setlength{\fboxsep}{1.2pt}\colorbox[HTML]{#2}{\textcolor{#1}{\textit{\texttt{\##2}}}}\endgroup}
\DeclareRobustCommand{\colorterm}[3][black]{\begingroup\setlength{\fboxsep}{1.2pt}\colorbox[HTML]{#2}{\textcolor{#1}{\textit{#3}}}\endgroup}
\DeclareRobustCommand{\colorprompt}[2][black]{\texttt{<color>}\hexcolor[#1]{#2}{\texttt{</color>}}}
\newcommand{\cmark}{\checkmark}
\newcommand{\xmark}{$\times$}

\usepackage{caption}
\usepackage{graphicx}
\usepackage{wrapfig}
\usepackage{booktabs}
\graphicspath{{common/}}

\newcommand{\PAtabwidth}{\linewidth}
\renewcommand{\PAdataexamples}{ Figure~\ref{fig:data-examples-main} shows examples from both streams.}

\title{\ours{}: Unified Any-Color Control for Image Generation and Editing}

\author[1,2]{Ji Xie}
\author[1,2]{Dewei Zhou}
\author[1]{Xinyu Huang}
\author[1,3]{Zhennan Chen}
\author[1]{Xun Wang}

\affiliation[1]{ByteDance Seed}
\affiliation[2]{Zhejiang University}
\affiliation[3]{Nanjing University}

\iflatexml\else
  \abstract{\exhyphenpenalty=10000
Any-color control, specifying an object's color by any 24-bit hex value instead of a coarse color word, is a practical requirement for image generation and editing in design work. Prior work often relies on dedicated color representations or specialized inference procedures, which can be hard to scale across colors or transfer to new architectures. In this paper, we present \ours{}, a single model that reads hex values directly from the prompt for both generation and editing. Our key insight is that the text encoders of recent image generators are language models that already associate hex values with colors, so a generator can learn hex control by grounding these values in object pixels. To this end, we construct \textbf{Paint-500K}, which labels object colors in real images and adds editing pairs, and combine it with \textbf{pure-color anchors}, solid-color images applied only at high-noise timesteps, for lightweight RGB alignment. We further introduce \textbf{Any Color Benchmark (ACBench)}, with \textbf{ACBench-T2I} and \textbf{ACBench-Edit}, to measure object-level hex color fidelity. Experiments and ablations show that \ours{} substantially improves any-color generation and editing over the base model, suggesting that arbitrary hex-color control can be learned as a native prompt-following behavior of modern image models.
}
\fi

\date{\today}

\hypersetup{
  pdftitle={Paint-Anything: Unified Any-Color Control for Image Generation and Editing},
  pdfauthor={Ji Xie, Dewei Zhou, Xinyu Huang, Zhennan Chen, Xun Wang},
  pdfsubject={Any-color control for image generation and editing},
  pdfkeywords={color control, text-to-image generation, image editing, hex color}
}

\begin{document}
\maketitle
\iflatexml
  \begin{abstract}
    \exhyphenpenalty=10000
Any-color control, specifying an object's color by any 24-bit hex value instead of a coarse color word, is a practical requirement for image generation and editing in design work. Prior work often relies on dedicated color representations or specialized inference procedures, which can be hard to scale across colors or transfer to new architectures. In this paper, we present \ours{}, a single model that reads hex values directly from the prompt for both generation and editing. Our key insight is that the text encoders of recent image generators are language models that already associate hex values with colors, so a generator can learn hex control by grounding these values in object pixels. To this end, we construct \textbf{Paint-500K}, which labels object colors in real images and adds editing pairs, and combine it with \textbf{pure-color anchors}, solid-color images applied only at high-noise timesteps, for lightweight RGB alignment. We further introduce \textbf{Any Color Benchmark (ACBench)}, with \textbf{ACBench-T2I} and \textbf{ACBench-Edit}, to measure object-level hex color fidelity. Experiments and ablations show that \ours{} substantially improves any-color generation and editing over the base model, suggesting that arbitrary hex-color control can be learned as a native prompt-following behavior of modern image models.

  \end{abstract}
\fi

\begin{figure}[t]
\centering
\includegraphics[width=0.90\textwidth]{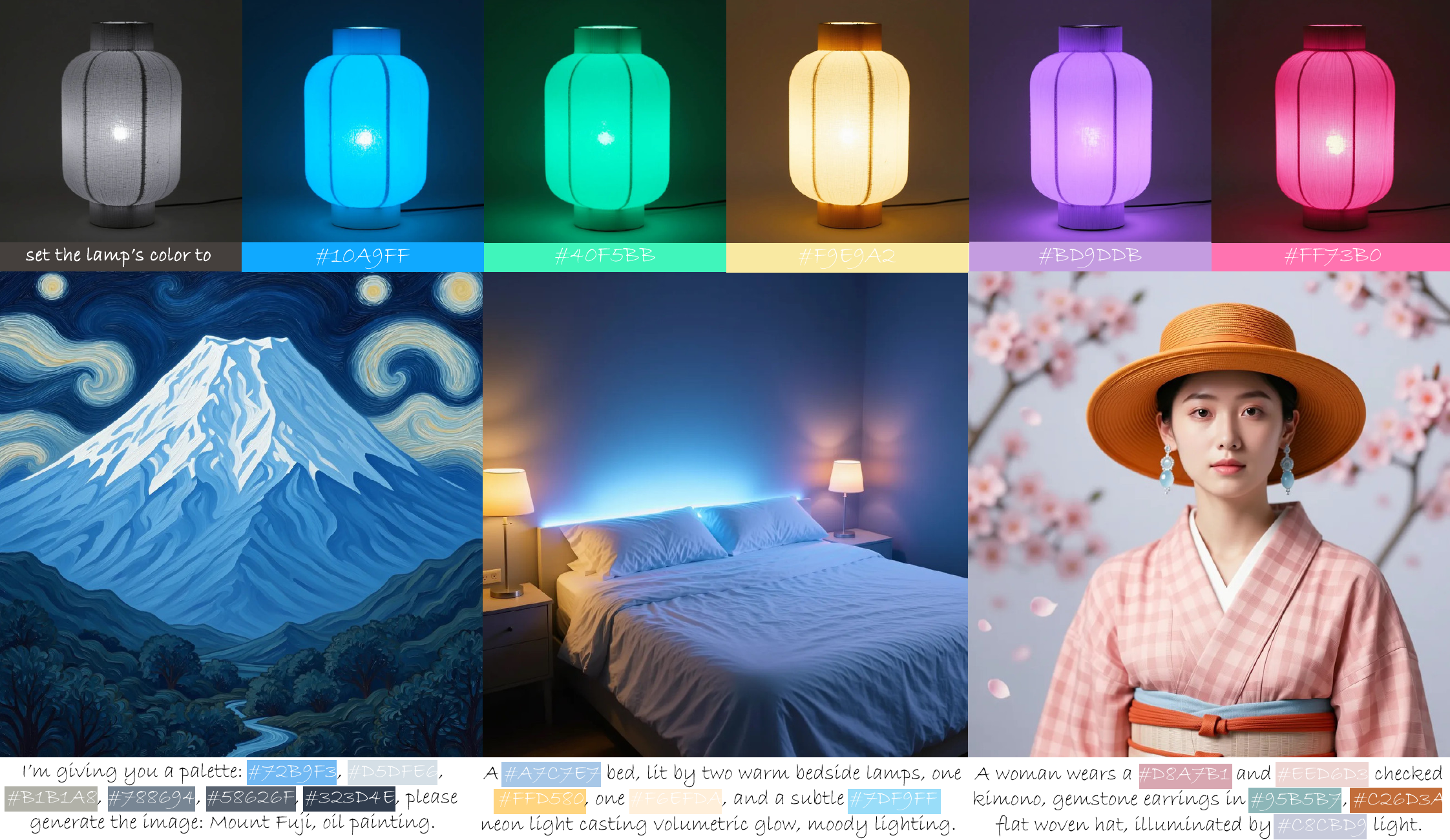}
\caption{\textbf{One hex-prompt interface for generation and editing.} A single finetuned model supports object recoloring and hex-conditioned generation through explicit color values in the prompt.}
\label{fig:teaser}
\end{figure}

\section{Introduction}
\label{sec:intro}

\textbf{How far is AI from a designer?} Today's image models can imagine rich scenes and render them in remarkable detail~\citep{podell2023sdxl,esser2024scaling,wu2025qwenimage,deng2025bagel,zhou2026metapoint,zhou20243dis,xie2026reconstruction,zhang2026images,fang2026unicorn,ummreflection2026,chen2026l2p,li2026videococo,zhou2026omnishow,zhou2024magictailor,jin2025srum,diao2026sensenova,liao2025imagegen}. Yet professional design~\citep{ye2026editable,ye2026genclaw,zhou2025dreamrenderer,du2025textcrafter,xu2026contextgen,li2026coco,liu2026hifi,liu2026ultratext,chen2025ragd} requires more than a convincing image: it requires following a color specification. A brand designer needs a logo in the brand's signature blue, and an online retailer needs product images in the specified red of a new collection. Neither requirement is captured by asking for ``blue'' or ``red'': both call for \textbf{explicit hex values}. Users should be able to choose an object's color as directly as a painter chooses colors for a canvas. We call this capability \emph{any-color control}: users specify target colors through 24-bit hex values.

Prior work has explored various approaches to color control. Some methods introduce task-specific modules or learned color tokens, while others use training-free inference-time techniques such as sampling guidance or attention/value manipulation~\citep{butt2024colorpeel,laria2025colorwave,shukla2024tintin,lobashev2025swguidance,shum2025coloralignment,qiu2025paletteguidance,aharoni2025palettealigned,yin2025colorctrl}. These designs can be difficult to scale across colors, hard to transfer to new architectures, or costly at inference. As a result, any-color generation, colorization, and editing are often treated as separate problems rather than as one native prompt-following capability.

\begin{wrapfigure}{r}{0.46\columnwidth}
\vspace{-0.8em}
\centering
\includegraphics[width=0.98\linewidth]{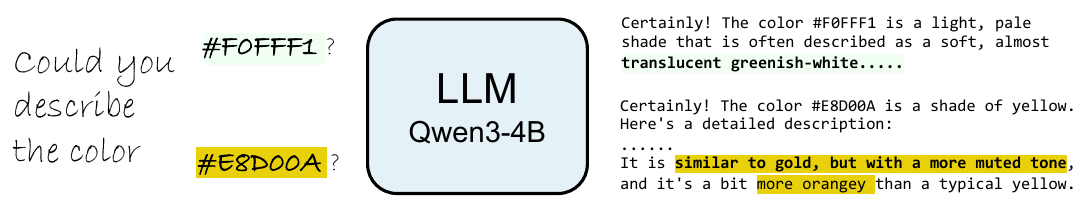}
\caption{\textbf{Hex color strings are legible to modern language models.} Even a compact 4B language model such as Qwen3-4B can associate raw hex codes such as \hexcolor{F0FFF1} and \hexcolor{E8D00A} with plausible color semantics, suggesting that hex strings can serve as a unified text interface for any-color generation and editing.}
\label{fig:llm-motivation}
\vspace{-0.6em}
\end{wrapfigure}

Fortunately, advances in large language models make hexadecimal color prompting a promising interface for this goal. A 24-bit RGB value can be written directly in text and bound to an object phrase, making it possible for a single model to support different any-color generation tasks with the same prompt language. As a simple motivating probe, Figure~\ref{fig:llm-motivation} shows that even a small LLM, Qwen3-4B~\citep{qwen2025qwen3}, can parse raw hex strings such as \hexcolor{F0FFF1} and \hexcolor{E8D00A} into plausible color descriptions. This suggests that hex strings can be used as a prompt-native numeric color interface.

\begin{figure}[t]
\centering
\includegraphics[width=\textwidth]{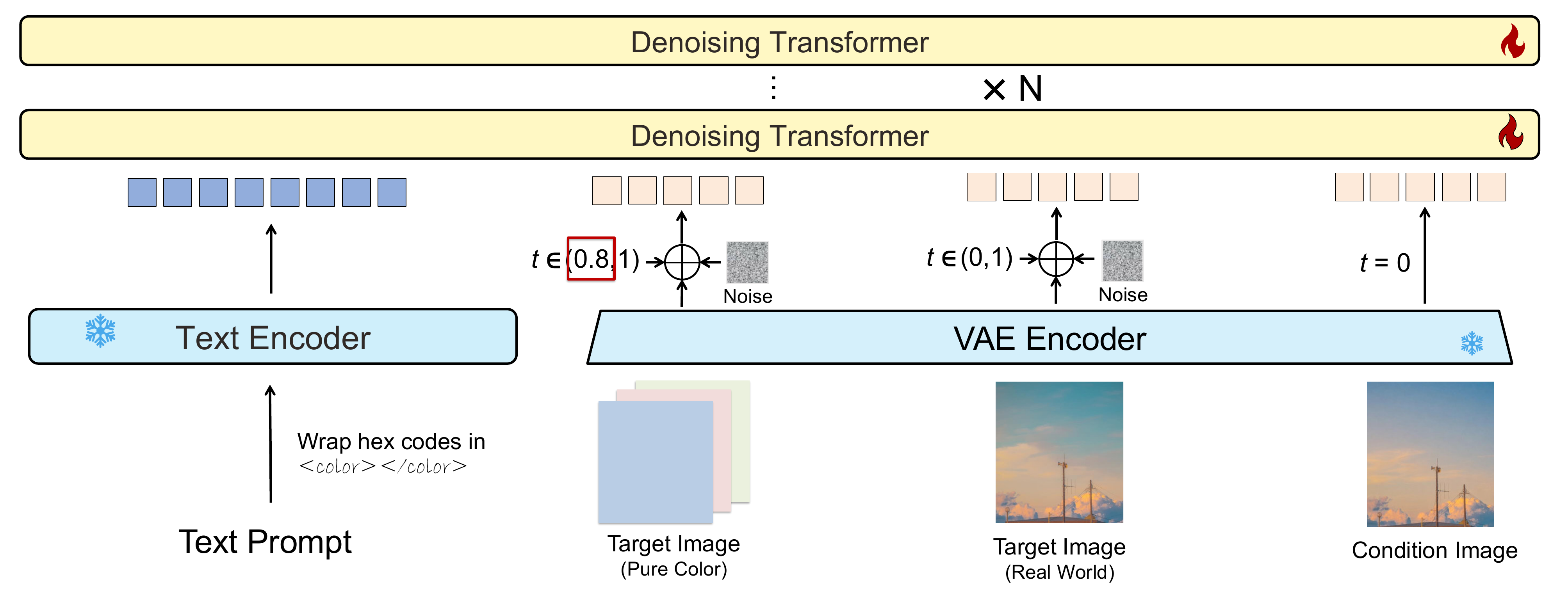}
\caption{\textbf{Training architecture.} Hex colors are represented directly in the text prompt and encoded by the text encoder as text tokens. The VAE encodes target images, and the denoising transformer is trained with the flow-matching objective. Pure-color targets are used only at high-noise timesteps $t\in[t_{\mathrm{gate}},1]$, real-world targets are sampled across $t\in(0,1)$, and optional condition images for editing are encoded as clean image latents at $t=0$.}
\label{fig:training-architecture}
\vspace{-1.0em}
\end{figure}

In this work, we present \ours{}, a unified model for any-color controllable generation and editing. \ours{} keeps color specification entirely inside the text prompt using explicit spans such as \colorprompt{AABBCC}, so that one interface covers both generation and editing. Instead of adding a separate inference-time color controller, we construct \textbf{Paint-500K} through a filtered data pipeline that turns real-world images into object-level hex-labeled supervision. VLM grounding identifies caption-relevant objects, segmentation masks isolate object pixels, and MeanShift clustering in CIELAB space~\citep{cie2018colorimetry} estimates dominant colors under natural illumination. Shadows make these real-image labels \emph{approximate colors}. To supply a clean low-level color reference, we additionally train on pure-color images as a \textbf{``pure-color anchor''}: each sample pairs a hex code with a solid-color target, and these samples are activated only at high-noise timesteps. This anchor gives the model a clean mapping from numeric hex strings to RGB statistics while low-noise training uses natural images.\PAintrofigs

To better evaluate any-color generation and editing, we introduce \textbf{Any Color Benchmark (ACBench)}\PAbenchnote{}, an object-level color-fidelity benchmark with two components: \textbf{ACBench-T2I} and \textbf{ACBench-Edit}. It measures the color fidelity of generated or edited object regions to the requested hex values. On \texttt{FLUX.2-4B}, \ours{} improves ACBench color-fidelity scores by \textbf{85.3\%} for generation and \textbf{28.3\%} for editing relative to the base model, and ablations show that pure-color anchors are an important ingredient in learning reliable pixel-space color control. Beyond ACBench, the same finetuned model achieves the highest average CompColor score~\citep{shomerchai2025colorbind} among the compared methods under both named-color and hex prompts. The gains are further supported by GenColorBench NCU evaluation and a human preference study (Appendices~\ref{app:benchmark-relation} and~\ref{app:user-study}).

Our contributions are:
\begin{list}{$\bullet$}{%
    \setlength{\leftmargin}{1.2em}%
    \setlength{\labelwidth}{0.8em}%
    \setlength{\labelsep}{0.4em}%
    \setlength{\itemsep}{0.2em}%
    \setlength{\topsep}{0.2em}%
}
    \item We discover that object-level hex supervision enables unified, prompt-native color control across generation and editing, and strengthen this capability with timestep-gated pure-color grounding.
    \item We develop a data pipeline that converts real images into object-level hex supervision for generation and editing. We use this pipeline to build Paint-500K.
    \item We introduce ACBench to evaluate object-level hex color fidelity in generation and editing, and demonstrate consistent gains across backbones and independent benchmarks.
\end{list}

\section{Related Work}
\label{sec:related}

\subsection{Color control in text-to-image generation}

Recent work has begun to make color a first-class control signal in text-to-image generation. Some methods learn or modify color representations, such as learnable color prompts~\citep{butt2024colorpeel}, CIELAB-aligned text embeddings~\citep{tsai2025colormecorrectly}, or numeric-color encoders for RGB/hex strings~\citep{butt2026numcolor}. Others use training-free inference-time techniques, such as sampling guidance, attention/value manipulation, or color-alignment objectives~\citep{shukla2024tintin,shum2025coloralignment,lobashev2025swguidance,agarwal2024colorstyle,laria2025colorwave,qiu2025paletteguidance,aharoni2025palettealigned}. These approaches improve color controllability, but often rely on additional control pathways, expensive inference-time optimization, or task-specific assumptions, which can limit unified task coverage.

NumColor~\citep{butt2026numcolor} learns numerical color embeddings, while BBQ-to-Image~\citep{kachlon2026bbq} uses structured prompts containing bounding boxes and RGB triplets.\footnote{Public inference code and checkpoints needed to run these methods on ACBench were unavailable at evaluation time. We nevertheless compare with NumColor using its reported GenColorBench NCU results in Appendix~\ref{app:benchmark-relation}.} Our model learns hex-to-color grounding through object-level supervision and uses hex codes directly in object descriptions and editing instructions. It unifies generation and editing without a dedicated color encoder or an intermediate layout representation containing bounding-box coordinates.

\subsection{Color editing and image colorization}

General image-editing systems based on prompt editing, instruction tuning, inversion, in-context editing, flow-based editing, or conditional control~\citep{hertz2023prompt,brooks2023instructpix2pix,kawar2022imagic,zhang2025context,kulikov2025flowedit,jiao2025unieditflow,zhang2023controlnet,xu2023inversion,xu2023cyclenet,zhou2026refineanything,xu2026phyedit} can change visual attributes. Here, we focus on edits specified by numerical color values. More specialized color-editing methods improve language-guided, region-aware, or continuous object color control~\citep{wang2023langrecol,dong2024cfnet,yin2024coloredit,yin2025colorctrl,yang2025continuouscolor,yang2026colourcrafter}. Palette- and histogram-based recoloring methods~\citep{chang2015palette,chao2023colorfulcurves,afifi2021histogan} provide explicit controls over image colors, rather than binding hex specifications to objects through text prompts. Image-reference conditioning methods such as IP-Adapter~\citep{ye2023ipadapter} provide continuous visual cues, but global image conditioning can entangle target color with style or texture. In parallel, image colorization methods~\citep{bahng2018text2colors,chang2023lcoins,chang2023lcad,li2024cocolc,liang2024ctrlcolor,an2025mtcolor,cong2024automatic} use masks, palette images, or semantic cues to constrain chromatic output. Our goal is to make the same hex-conditioned prompt interface work across generation and editing with a single finetuned model.

\begin{figure}[t]
\centering
\includegraphics[width=\textwidth]{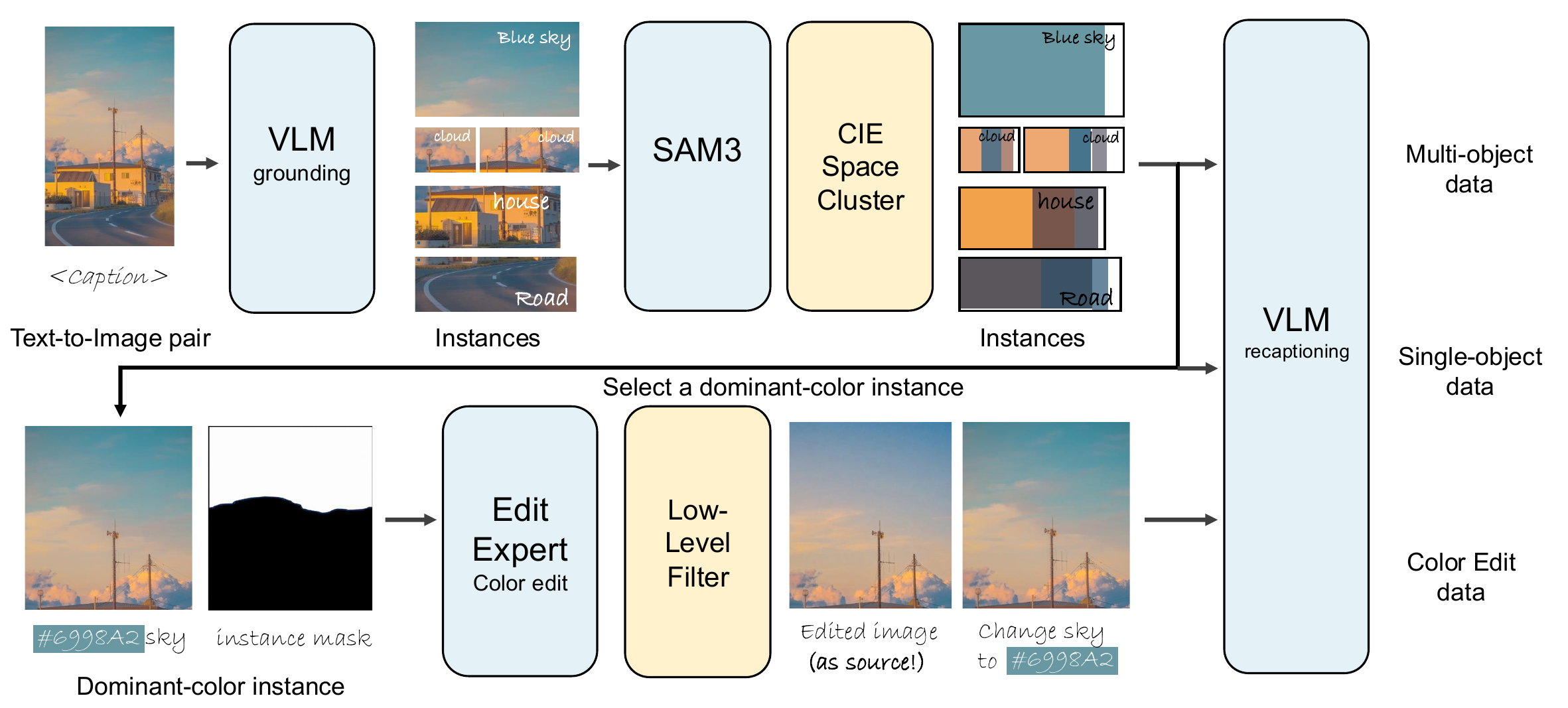}
\caption{\textbf{Paint-500K data pipeline.} Real image-caption pairs are converted into hex-conditioned T2I and editing data through VLM grounding, SAM3 masks, CIE-space color extraction, edit synthesis, filtering, and recaptioning.}
\label{fig:data-pipeline}
\end{figure}

\section{Method}
\label{sec:method}

\subsection{Preliminaries}
\label{sec:method-prelim}
We build on a rectified-flow text-to-image model~\citep{lipman2023flowmatching,esser2024scaling} with a text encoder, a VAE~\citep{kingma2014autoencoding,rombach2022ldm}, and a denoising transformer~\citep{esser2024scaling}. Let $y$ denote the text prompt and $I^\star$ the target image. For editing, $I^c$ denotes the source image. The frozen text encoder maps $y$ into embeddings $\mathbf{c}$. The VAE encoder maps the target image into a latent sequence $\mathbf{z}_0=E_{\mathrm{vae}}(I^\star)$, and for editing the same VAE encodes the optional source image into a clean condition latent $\mathbf{z}^c=E_{\mathrm{vae}}(I^c)$.

Training follows the flow-matching objective. We sample $\epsilon\sim\mathcal{N}(0,\mathbf{I})$ and $t\sim\mathcal{U}(0,1)$, where $t=0$ corresponds to the clean target latent and $t=1$ corresponds to pure noise, and form
\begin{equation}
\mathbf{z}_t=(1-t)\mathbf{z}_0+t\epsilon,\qquad
\mathbf{v}^{\star}=\epsilon-\mathbf{z}_0.
\label{eq:flow-target}
\end{equation}
The denoiser predicts a velocity from the noisy target latent, the timestep, the text tokens, and optional image-condition tokens:
\begin{equation}
\widehat{\mathbf{v}}_{\theta}
=D_{\theta}\!\left([\mathbf{z}_t;\mathbf{z}^c],t,\mathbf{c}\right),
\label{eq:flux-forward}
\end{equation}
where $[\mathbf{z}_t;\mathbf{z}^c]$ denotes sequence concatenation and reduces to $\mathbf{z}_t$ for T2I samples. The base loss is
\begin{equation}
\mathcal{L}_{\mathrm{FM}}
=\mathbb{E}_{\mathbf{z}_0,\epsilon,t,y}\left[
\left\|\widehat{\mathbf{v}}_{\theta}-\mathbf{v}^{\star}\right\|_2^2
\right].
\label{eq:flow-loss}
\end{equation}
Thus generation and editing share the same denoising objective and differ only in whether the clean condition-image latent is concatenated to the noisy target latent.

\subsection{Training pipeline}
\label{sec:method-training}
Figure~\ref{fig:training-architecture} summarizes our training pipeline for grounding a shared hex-prompt interface through object-level supervision and timestep-gated pure-color anchors. The model follows ordinary text prompts for generation and takes a source image for editing.

\textbf{Explicit color-tag wrapping.}
We mark each 24-bit sRGB hex specification as a color attribute by enclosing it in \texttt{<color>} and \texttt{</color>} tags directly in the prompt. Because the text encoder remains frozen during finetuning, we make the color span explicit in its input before encoding. For example, prompts can write ``a photo of a \colorprompt{CFEFFB} colored car'' or ``change the bag to \colorprompt{FFF8C4}.'' Empirically, this wrapping improves object-level color fidelity and compositional color binding, with gains on ACBench-T2I, ACBench-Edit, and CompColor (Table~\ref{tab:ablation-results}).

\textbf{Unified generation and editing format.}
We jointly train generation and editing so that one model learns color perception, object-color binding, and color-conditioned editing through the same hex-prompt interface. Every sample contains a target image and a hex-conditioned prompt, and each editing sample also contains a source image encoded as a clean condition-latent branch.

\textbf{Pure-color anchor supervision.}
Real images provide object semantics, but their measured colors are noisy: shadows can map the same object to many plausible RGB values, leaving only an approximate color. We therefore add \textbf{pure-color anchors}: we sample 24-bit RGB values, render each value as a solid-color image, and pair it with a prompt containing the corresponding \texttt{<color>\#HEX</color>} token. These anchors provide a clean low-level signal for hex-to-RGB grounding, so we use them to focus training on color-token alignment rather than on object semantics. Motivated by early color stabilization in pure-color generation trajectories (Appendix~\ref{app:timestep-vis}), we apply a \textbf{high-noise gate}: pure-color samples are trained only with $t\sim\mathcal{U}(t_{\mathrm{gate}},1.0)$, while real-image T2I and editing samples use $t\sim\mathcal{U}(0,1)$. For $t_{\mathrm{gate}}=1$, pure-color samples are trained at $t=1$. Table~\ref{tab:gate-results} shows that high-noise gating improves generation and editing over ungated anchors.

\textbf{Training loss.}
The final training objective combines the three streams:
\begin{equation}
\mathcal{L}
=\lambda_{\mathrm{t2i}}\mathcal{L}_{\mathrm{t2i}}
+\lambda_{\mathrm{edit}}\mathcal{L}_{\mathrm{edit}}
+\lambda_{\mathrm{rgb}}\mathcal{L}_{\mathrm{rgb}},
\label{eq:final-loss}
\end{equation}
where each term uses the flow-matching loss in Eq.~\ref{eq:flow-loss}. $\mathcal{L}_{\mathrm{t2i}}$ is computed on real-image color-generation samples, $\mathcal{L}_{\mathrm{edit}}$ on paired image-editing samples, and $\mathcal{L}_{\mathrm{rgb}}$ on pure-color anchor samples under the high-noise gate. We finetune the denoising transformer with the VAE and text encoder frozen. Additional mixture and optimization details are provided in Appendix~\ref{app:training-data}.


\subsection{Data pipeline}
\label{sec:method-data}
\begin{wrapfigure}{l}{0.48\columnwidth}
\vspace{-0.8em}
\centering
\includegraphics[width=0.98\linewidth]{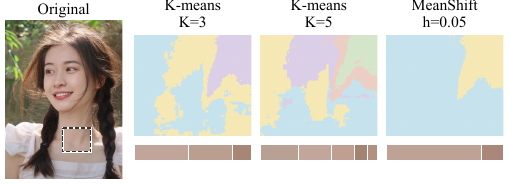}
\caption{\textbf{MeanShift reduces redundant color splits compared with fixed-$K$ $k$-means.} Both methods cluster the outlined skin patch in CIELAB before filtering. Maps show cluster membership, and bars show mean-RGB palettes weighted by pixel share.}
\label{fig:skin-clustering-main}
\vspace{-0.6em}
\end{wrapfigure}

\textbf{Extracting object colors.}
Extracting a representative color from an object is crucial for reliable color supervision: lighting and shadows can make a single-colored object contain many pixel values. A common approach is RGB-space clustering~\citep{yang2026colourcrafter,pylette2026,kachlon2026bbq}, typically using $k$-means. This has two limitations: (I) RGB distances do not reflect human perception~\citep{cie2018colorimetry,yang2026colourcrafter}, potentially separating perceptually similar colors or merging distinct ones, and (II) objects have varying numbers of dominant colors, so a fixed $K$ can split shading and noise into redundant labels. Inspired by ColorBind~\citep{shomerchai2025colorbind}, we use the perceptual CIELAB space~\citep{cie2018colorimetry}. We further introduce MeanShift~\citep{comaniciu2002meanshift} into object-color annotation to adapt the cluster count to each object's color distribution. Empirically, MeanShift reduces redundant color splits relative to fixed-$K$ $k$-means in CIELAB, yielding a more coherent dominant-color region (Figure~\ref{fig:skin-clustering-main}).


\begin{figure}[t]
\centering
\includegraphics{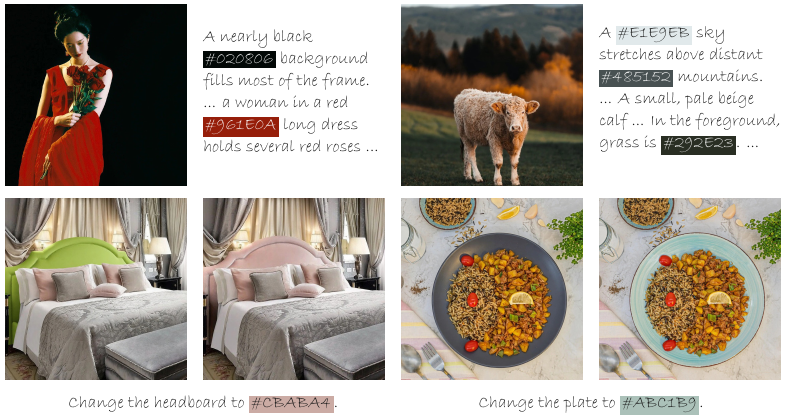}
\caption{\textbf{Paint-500K training examples.} Top: T2I samples whose captions carry the target hex colors (shortened with ellipses). Bottom: editing pairs, each with the source (left), the target (right) and the instruction. Further examples are in Appendix~\ref{app:data-samples}.}
\label{fig:data-examples-main}
\end{figure}

\textbf{Binding colors to objects and captions.}
Paint-500K uses high-quality real image-caption pairs from an internal collection. A VLM identifies caption-relevant objects and their bounding boxes, and SAM3~\citep{carion2025sam3} produces the object masks used for color extraction. We filter clusters by pixel coverage and discard instances without sufficient retained coverage. The largest retained cluster provides an sRGB hex label for its object. The VLM then rewrites the caption to bind each label to the corresponding noun phrase using explicit \texttt{<color>\#HEX</color>} spans. Figure~\ref{fig:data-pipeline} shows the full pipeline. Its filtering details and VLM templates are given in Appendices~\ref{app:training-data} and~\ref{app:prompt-templates}.

\textbf{Generation and editing streams.}
For T2I supervision, multi-object captions teach compositional color binding, while single-object crops emphasize direct object-color perception. We collect 400K T2I samples: 100K single-object and 300K multi-object examples. For editing, a pretrained editing model (Appendix~\ref{app:training-data}) recolors each single-object image to form the source, and the original photograph serves as the target, paired with an instruction to restore its hex color. Only the source is generated, so the targets carry no generative artifacts. Filtering for scene preservation, meaningful recoloring, and instruction consistency yields 100K editing samples that use the same prompt-native color syntax.\PAdataexamples

\begin{figure}[t]
    \centering
    \includegraphics[width=0.96\textwidth]{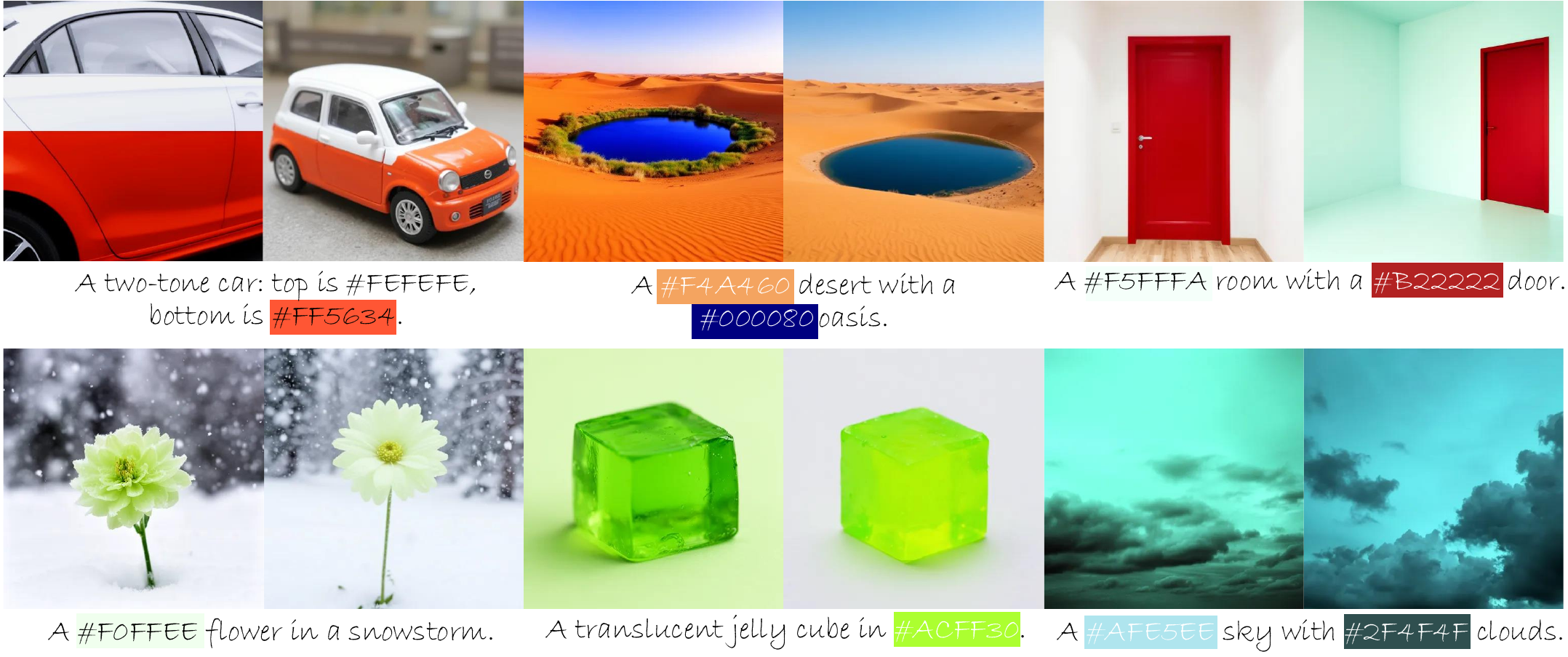}
    \caption{\textbf{Qualitative comparison with the base model.} Each pair shows \texttt{FLUX.2-4B} (left) and \ours{} (right) on the same hex-conditioned prompt. The baseline often produces colors that are semantically plausible but numerically distant from the requested hex value.}
    \label{fig:qualitative-vs-baseline}
    \vspace{-0.5em}
\end{figure}

\section{Experiments}
\label{sec:experiments}

\subsection{Benchmarks and metrics}
\paragraph{CompColor benchmark.}
CompColor~\citep{shomerchai2025colorbind} is a compositional color-binding benchmark that measures whether a generator can faithfully bind two distinct colors to two objects in the same prompt. Each prompt has the form ``a \{color\} colored \{object\} and a \{color\} colored \{object\}'', with color words drawn from a fixed \emph{named-color} palette, e.g., ``a \colorterm{AFEEEE}{paleturquoise} colored shirt and a \colorterm{FFFFF0}{ivory} colored bench''. The released baseline table covers a wide range of color-binding methods evaluated under this named-color setting. Our task instead specifies prompt colors as 24-bit hex strings. We therefore evaluate CompColor under an additional hex-translated transfer protocol that replaces each color word with its canonical hex value while keeping the original object compositions intact, e.g., ``a \hexcolor{AFEEEE} colored shirt and a \hexcolor{FFFFF0} colored bench''. More details about this benchmark and our hex-translated protocol are in Appendix~\ref{app:compcolor-transfer}.

\paragraph{Any Color Benchmark (ACBench).}
To evaluate object-level color fidelity under 24-bit hex prompts in both generation and editing, we introduce \textbf{ACBench}. \emph{ACBench-T2I} contains 1000 generation prompts over common object categories, and \emph{ACBench-Edit} contains 500 real-image recoloring prompts. Following prior object-based benchmarks~\citep{ghosh2023geneval}, we pair randomly sampled hex colors with everyday objects, animals, and plants. ACBench covers three settings:

\begin{list}{$\bullet$}{%
    \setlength{\leftmargin}{1.2em}%
    \setlength{\labelwidth}{0.8em}%
    \setlength{\labelsep}{0.4em}%
    \setlength{\itemsep}{0.1em}%
    \setlength{\topsep}{0.1em}%
}
    \item \textbf{Single-object (Single)} (500 prompts): one object with one target color, e.g., ``a photo of a \hexcolor{CFEFFB} car.''
    \item \textbf{Two-object (Two)} (500 prompts): two objects, each assigned a distinct target color, e.g., ``a photo of a \hexcolor{FFF8C4} dog and a \hexcolor{CFEFFB} chair.''
    \item \textbf{Edit} (500 prompts): one object with one target color, e.g., ``change the bag to \hexcolor{CFEFFB}.''
\end{list}


\paragraph{Protocol and metric.} We use SAM3~\citep{carion2025sam3} to obtain the target masks. For T2I, we segment the prompted object in the generated image. For editing, we segment the target object in the source image, inspect the mask manually, and reuse the same source mask across the compared edited outputs. For a target mask $M$, let $\bar{\mathbf{c}}=|M|^{-1}\sum_{p\in M}I(p)$ be the mean sRGB vector on the 0 to 255 scale. Following prior work~\citep{shomerchai2025colorbind,butt2025gencolorbench}, we compare the estimated object color with the target. Given target RGB $\mathbf{c}^{\star}$, we compute $\mathrm{MAE}=\frac{1}{3}\sum_{k\in\{R,G,B\}}|\bar{c}_k-c_k^{\star}|$, and failed localization receives a score of 0. CIELAB estimators and CIEDE2000 evaluation appear in Appendices~\ref{app:acbench-reliability} and~\ref{app:benchmark-relation}.

We convert MAE into a normalized score using a piecewise linear function:
\begin{equation}
    s = 100 \cdot \max\!\left(0,\; 1 - \frac{\max(0,\, \mathrm{MAE} - 16)}{48}\right).
\end{equation}
The score measures region-level color fidelity while allowing natural appearance variation. MAE of at most 16, measured as the average absolute channel error, receives full score. MAE between 16 and 64 is linearly penalized, and MAE of at least 64 receives 0. The linear interval retains graded credit for imperfect color matches, preserving distinctions that a tighter cutoff would collapse to zero. We generate one image per T2I prompt with one seed per prompt and report scores separately for the equally sized Single and Two splits, with Overall as their arithmetic mean. Beyond these automatic scores, we also conduct a user study to examine whether people prefer our color results to those of the base model (Appendix~\ref{app:user-study}).

\subsection{Experimental setup}

\begin{table}[!t]
    \centering
    \footnotesize
    \setlength{\abovecaptionskip}{6pt}
    \setlength{\tabcolsep}{3.5pt}
    \renewcommand{\arraystretch}{1.15}
    \begin{tabular*}{\linewidth}{@{\extracolsep{\fill}}lcccccccc@{}}
\toprule
& & \multicolumn{3}{c}{ACBench-T2I $\uparrow$} & ACBench-Edit $\uparrow$ & \multicolumn{3}{c}{CompColor $\uparrow$} \\
\cmidrule(lr){3-5} \cmidrule(lr){6-6} \cmidrule(lr){7-9}
Model & \shortstack{Total\\params.} & Single & Two & Overall & Edit & Single & Close & Distant \\
\midrule
\texttt{SD1.5}                       & 1.0B  & 15.00 & 8.00 & 11.50 & ---            & 0.53$^{\dagger}$          & 0.36$^{\dagger}$          & 0.30$^{\dagger}$          \\
\texttt{FLUX.1}                          & 17B   & 25.00 & 21.00 & 23.00 & ---            & 0.56$^{\dagger}$          & 0.54$^{\dagger}$          & 0.49$^{\dagger}$          \\
\texttt{Z-Image-Turbo}                                                  & 10B   & 32.11 & 35.00 & 33.56 & ---            & 0.65          & 0.69          & 0.66          \\
\texttt{Qwen-Image\,/\,Edit}                                       & 27B   & 26.07 & 24.34 & 25.21 & 54.40          & 0.59          & 0.59          & 0.63          \\
\texttt{FLUX.2-9B}                                             & 17B   & 36.24 & 40.07 & 38.15 & 64.53          & 0.75          & 0.67          & 0.68          \\
\texttt{FLUX.2-dev}                                                & 56B   & 51.32 & 52.08 & 51.70 & 68.87          & 0.73          & 0.72          & \textbf{0.79}          \\
\texttt{FLUX.2-4B} & \multirow{2}{*}{8B} & \multirow{2}{*}{37.45} & \multirow{2}{*}{36.60} & \multirow{2}{*}{37.02} & \multirow{2}{*}{58.90} & 0.74          & 0.70          & 0.73          \\
\texttt{FLUX.2-4B} (Hex) & & & & & & 0.34          & 0.41          & 0.38          \\
\texttt{Z-Image} Base                                                   & 10B   & 32.15 & 34.74 & 33.45 & ---   & ---           & ---           & ---           \\
\midrule
\multicolumn{9}{l}{\emph{Color-specialized methods}} \\
\texttt{ColorBind/Edit}                                    & var.         & 30.42 & 34.17 & 32.30 & 60.38 & 0.72          & 0.71          & 0.73          \\
\texttt{CtrlColor}                                              & SD1.5-based  & ---   & ---   & ---   & 57.46 & ---           & ---           & ---           \\
\texttt{ColorPeel}                                               & SD1.4-based  & 45.28 & 34.63 & 39.96 & ---   & 0.68          & 0.62          & 0.64          \\
\texttt{ColorWave}\textsuperscript{*}                           & SDXL-based   & 50.36 & 42.71 & 46.54 & ---   & 0.72          & 0.68          & 0.70          \\
\midrule
\texttt{FLUX.2-4B} + \textbf{Ours} & \multirow{2}{*}{8B} & \multirow{2}{*}{\textbf{72.67}} & \multirow{2}{*}{\textbf{64.49}} & \multirow{2}{*}{\textbf{68.58}} & \multirow{2}{*}{\textbf{75.57}} & \textbf{0.81} & \textbf{0.80} & 0.76          \\
\texttt{FLUX.2-4B} + \textbf{Ours} (Hex) & & & & & & 0.78          & \textbf{0.80} & \textbf{0.79} \\
\texttt{Z-Image} Base + \textbf{Ours}                                                      & 10B   & 56.76 & 50.78 & 53.77 & ---   & 0.77 & 0.77 & 0.76 \\
\bottomrule
\end{tabular*}

    \caption{Main results: ACBench (0--100) and CompColor (0--1). CompColor uses named colors unless marked \emph{(Hex)}, and paired rows share ACBench results. ---: unreported. $\dagger$: quoted from ColorBind~\citep{shomerchai2025colorbind}. \textsuperscript{*}: our reproduction. Reruns use official defaults (FLUX.2-4B Edit CFG 2.0). NCU and additional CompColor results: Appendices~\ref{app:benchmark-relation} and~\ref{app:compcolor-transfer}.}
    \label{tab:main-results}
    \vspace{-1.0em}
\end{table}

\textbf{Baseline and implementation details.} We finetune \texttt{FLUX.2-klein-base-4B}~\citep{bfl2026flux2klein4bmodel} and \texttt{Z-Image} Base~\citep{zimage2025} with the same training recipe. Both models are trained for 4000 steps on 4 GPUs, using Adam with a global batch size of 72 and a learning rate of $2\times10^{-5}$. Throughout the paper, \texttt{FLUX.2-4B} and \texttt{FLUX.2-9B} denote the undistilled klein base checkpoints.\footnote{The ``4B'' and ``9B'' suffixes refer to the DiT parameter count, whereas the Total params. column in Table~\ref{tab:main-results} reports the total parameter count, including the bundled Qwen3 text encoder.} The FLUX ablations use the 4B backbone with the same optimization settings.

\textbf{Evaluation details.} Table~\ref{tab:main-results} compares complete systems, whereas same-backbone ablations assess our training recipe. The open-source baselines are SD1.5~\citep{rombach2022ldm}, FLUX.1~\citep{flux2023}, Z-Image-Turbo~\citep{zimage2025}, Qwen-Image and Qwen-Image-Edit~\citep{wu2025qwenimage}, FLUX.2-4B~\citep{bfl2026flux2klein4bmodel}, FLUX.2-9B~\citep{bfl2026flux2klein9b} and FLUX.2-dev~\citep{bfl2025flux2dev}. The color-specialized baselines are ColorBind/Edit~\citep{shomerchai2025colorbind}, CtrlColor~\citep{liang2024ctrlcolor}, ColorPeel~\citep{butt2024colorpeel} and ColorWave~\citep{laria2025colorwave}. NumColor inference code and checkpoints were unavailable at evaluation time, so we use its published NCU results (Appendix~\ref{app:benchmark-relation}). CompColor includes named-color and hex prompts (Appendix~\ref{app:compcolor-transfer}).


\begin{figure}[t]
\centering
\includegraphics[page=1,width=0.96\textwidth]{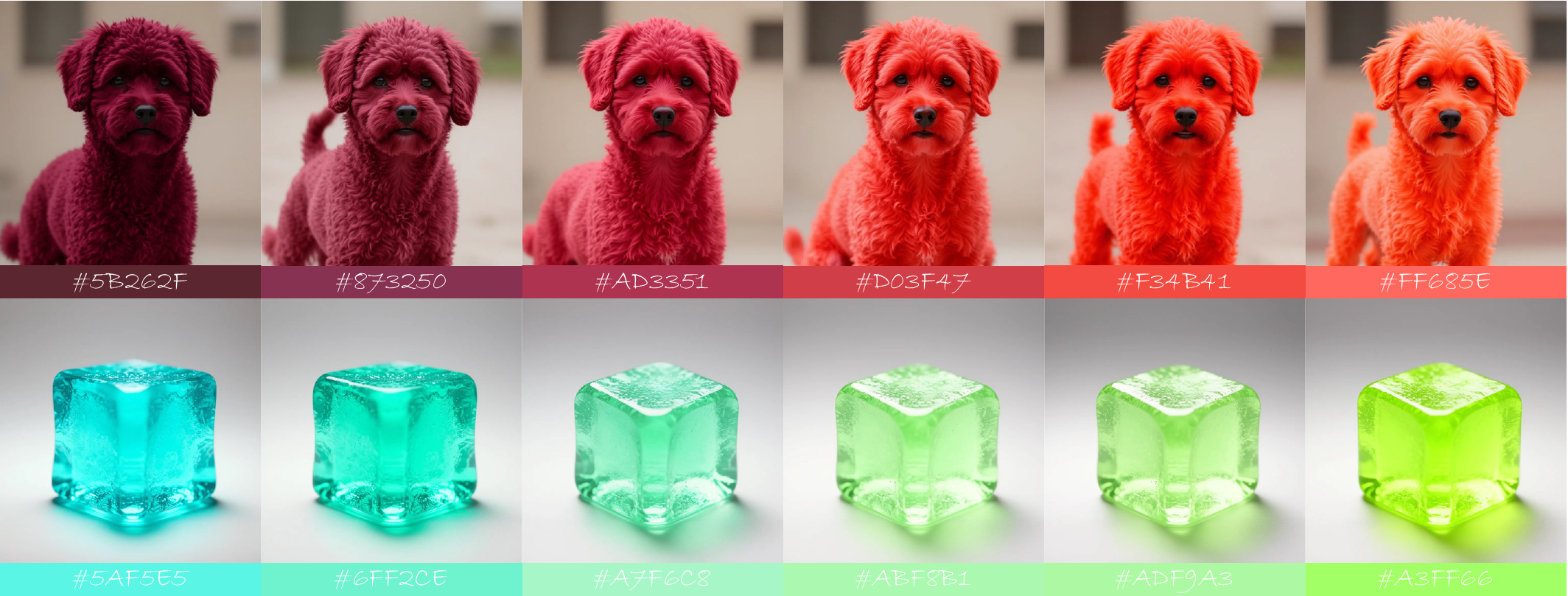}
\caption{\textbf{Hex-conditioned generation.} \ours{} follows a range of requested hex colors for the same object. All results use the same random seed.}
\label{fig:qualitative-generation}
\end{figure}

\begin{figure}[t]
\centering
\includegraphics[page=2,width=0.96\textwidth]{figures/qualitative_result.pdf}

\caption{\textbf{Hex-conditioned editing.} Given a source image and a hex-color instruction, \ours{} follows the requested hex color when recoloring the target object.}
\label{fig:qualitative-editing}
\end{figure}

\PAqualitativeEarly

\subsection{Quantitative results}
Table~\ref{tab:main-results} shows that our finetuned \texttt{FLUX.2-4B} improves over its base checkpoint across three complementary aspects of color control: object-level hex color fidelity for text-to-image generation on ACBench-T2I, recoloring fidelity for image editing on ACBench-Edit, and compositional color binding on CompColor.

\textbf{Hex supervision enables an 8B model to outperform a 56B model.} Within the FLUX.2 family, ACBench-T2I Overall and ACBench-Edit scores increase across the off-the-shelf models with 8B, 17B, and 56B total parameters. Yet our finetuned 8B model exceeds the 56B model by 16.88 points on ACBench-T2I and 6.70 points on ACBench-Edit. The same training recipe also produces a large gain on \texttt{Z-Image} Base.

\textbf{Finetuning closes the word-to-hex gap.} On CompColor, replacing color names with raw hex strings reduces the \texttt{FLUX.2-4B} base model's average score from 0.72 to 0.38. After localized hex supervision, the hex-prompt average more than doubles and exceeds the base model's named-color average. The named-color average also improves from 0.72 to 0.79, showing that finetuning preserves and improves the model's existing compositional color-binding ability.

\textbf{\ours{} outperforms the evaluated specialized systems.} On ACBench, \ours{} exceeds the strongest specialized baseline in each task: ColorWave by 22.04 points on ACBench-T2I and ColorBind/Edit by 15.19 points on ACBench-Edit. The independent GenColorBench NCU evaluation also ranks \ours{} highest among the compared methods, while a human preference study favors its color results over those of the base model (Appendices~\ref{app:benchmark-relation} and~\ref{app:user-study}).

\subsection{Ablation study}

\begin{wraptable}{r}{0.66\columnwidth}
\vspace{-0.8em}
\centering
\footnotesize
\setlength{\abovecaptionskip}{6pt}
\setlength{\tabcolsep}{3pt}
\renewcommand{\arraystretch}{1.12}
\renewcommand{\PAtabwidth}{0.64\columnwidth}
\begin{tabular*}{\PAtabwidth}{@{\extracolsep{\fill}}cccccccc@{}}
\toprule
Base & LoRA & Wrap & Pure & Gate & T2I $\uparrow$ & Edit $\uparrow$ & CompC. $\uparrow$ \\
\midrule
\cmark & --- & --- & --- & --- & 37.02 & 58.90 & 0.38 \\
\midrule
\xmark & \xmark & \xmark & \xmark & \xmark & 44.06 & 65.81 & 0.52 \\
\xmark & \xmark & \cmark & \cmark & \xmark & 63.83 & 70.39 & 0.71 \\
\xmark & \xmark & \cmark & \xmark & \xmark & 57.16 & 73.89 & 0.71 \\
\xmark & \xmark & \xmark & \cmark & \cmark & 48.01 & 64.08 & 0.59 \\
\xmark & \cmark & \cmark & \cmark & \cmark & 48.45 & 58.71 & 0.54 \\
\xmark & \xmark & \cmark & \cmark & \cmark & \textbf{68.58} & \textbf{75.57} & \textbf{0.79} \\
\bottomrule
\end{tabular*}

\caption{Module ablations on \texttt{FLUX.2-4B}. Base denotes the pretrained model without finetuning. For the other rows, LoRA $\times$ denotes full finetuning and $\checkmark$ denotes rank-256 LoRA. Pure denotes pure-color anchors. T2I and Edit are the ACBench scores, and CompC. is the hex-prompt CompColor average.}
\label{tab:ablation-results}
\end{wraptable}

We ablate four choices in the final recipe: color-token wrapping, pure-color anchors, high-noise gating, and full-model finetuning. Table~\ref{tab:ablation-results} shows that the complete recipe is best across ACBench generation, ACBench editing, and CompColor, while Table~\ref{tab:gate-results} isolates the gate threshold.

\begin{samepage}
\textbf{Full finetuning and LoRA.}
With the complete recipe, a shared learning rate and 4000 steps, full finetuning exceeds rank-256 LoRA: 68.58 vs. 48.45 (T2I), 75.57 vs. 58.71 (Edit), and 0.79 vs. 0.54 (CompColor).\par
\end{samepage}

\textbf{Wrapping clarifies hex syntax, while anchors provide a clean color reference.}
Without pure-color anchors, bare-hex finetuning reaches only 44.06 and 65.81 on ACBench-T2I and ACBench-Edit. Color-token wrapping raises these scores to 57.16 and 73.89, giving the model a clearer textual handle for raw hex strings. With wrapping enabled, adding pure-color anchors provides a clean hex-to-RGB reference. Without gating, however, generation improves to 63.83 while editing decreases to 70.39. The ungated anchor objective therefore helps generation at the cost of editing fidelity.

\begin{wraptable}{l}{0.58\columnwidth}
\vspace{-0.8em}
\centering
\footnotesize
\setlength{\abovecaptionskip}{6pt}
\setlength{\tabcolsep}{3.5pt}
\renewcommand{\arraystretch}{1.12}
\renewcommand{\PAtabwidth}{0.56\columnwidth}
\begin{tabular*}{\PAtabwidth}{@{\extracolsep{\fill}}lccccc@{}}
\toprule
$t_{\mathrm{gate}}$ & 0.0 & 0.7 & 0.8 & 0.9 & 1.0 \\
\midrule
ACBench-T2I $\uparrow$ & 63.83 & 65.93 & \textbf{68.58} & 67.54 & 67.62 \\
ACBench-Edit $\uparrow$ & 70.39 & 72.68 & \textbf{75.57} & 73.68 & 71.49 \\
CompColor $\uparrow$ & 0.71 & \textbf{0.79} & \textbf{0.79} & 0.77 & 0.76 \\
\bottomrule
\end{tabular*}

\caption{Timestep-gate ablation on \texttt{FLUX.2-4B}, with wrapping and anchors enabled. $t_{\mathrm{gate}}=0$ matches the ungated anchor row in Table~\ref{tab:ablation-results}.}
\label{tab:gate-results}
\end{wraptable}

\textbf{High-noise gating makes anchor supervision more effective.}
Restricting pure-color supervision to high-noise timesteps matches where color emerges in pure-color trajectories (Appendix~\ref{app:timestep-vis}) and leaves lower-noise training to real-image generation and editing. With wrapping enabled, our gated recipe improves ACBench-T2I by 4.75 points and ACBench-Edit by 5.18 points over ungated anchors, while also improving CompColor. It also outperforms training without anchors on both generation and editing. Thresholds 0.7, 0.8, and 0.9 all improve both tasks over ungated anchors (Table~\ref{tab:gate-results}), showing low sensitivity. We use $t_{\mathrm{gate}}=0.8$.

\subsection{Qualitative results}
\ours{} better matches the requested hex values across object categories and color families, whereas \texttt{FLUX.2-4B} often produces plausible but numerically inaccurate colors under the same prompts (Figure~\ref{fig:qualitative-vs-baseline}). Figure~\ref{fig:qualitative-generation} shows generation for the same object under different hex specifications, while Figure~\ref{fig:qualitative-editing} shows recoloring a target object in a source image according to a hex-color instruction.


\section{Limitations and Future Work}

\label{sec:limitations}
Our current training data do not include palette-specific supervision. Future work can add palette-specific supervision and extend the same unified interface to a broader range of color-control tasks.

\section{Conclusion}
\label{sec:conclusion}
We presented \ours{}, showing that pretrained image models can learn any-color control as a unified prompt-native capability for generation and editing. Paint-500K supplies object-level hex supervision through perceptual color clustering, with our empirical analysis motivating MeanShift over fixed-$K$ clustering. Overall, \ours{} improves ACBench color fidelity by 85.3\% for generation and 28.3\% for editing relative to the base model.

\FloatBarrier
\clearpage
{
\small
\addtolength{\bibsep}{-0.5pt}
\bibliographystyle{styles/ieeenat_fullname}
\bibliography{common/refs}
}

\clearpage
\appendix
\section*{Supplementary Material Overview}
We provide additional information in the supplementary material, as outlined below:
\begin{list}{$\bullet$}{%
    \setlength{\leftmargin}{1.2em}%
    \setlength{\labelwidth}{0.8em}%
    \setlength{\labelsep}{0.4em}%
    \setlength{\itemsep}{2pt}%
    \setlength{\topsep}{2pt}%
    \setlength{\parsep}{0pt}%
}
\item Sec.~\ref{app:benchmark-relation}: How ACBench relates to existing color benchmarks, with the GenColorBench NCU transfer check.
\item Sec.~\ref{app:benchmark-prompts}: ACBench construction, prompt sampling and baseline settings.
\item Sec.~\ref{app:compcolor-transfer}: The hex-translated CompColor protocol and the full original baseline roster.
\item Sec.~\ref{app:training-data}: Paint-500K composition, the training mixture and the tools used to build it.
\item Sec.~\ref{app:timestep-vis}: Pure-color generation trajectories motivating the high-noise gate.
\item Sec.~\ref{app:prompt-templates}: Prompt templates for grounding, filtering and recaptioning.
\item Sec.~\ref{app:evaluator-settings}: Evaluator settings used for every reported score.
\item Sec.~\ref{app:acbench-reliability}: ACBench reliability under alternative region statistics, SAM3 thresholds and localization failures.
\item Sec.~\ref{app:anchor-resolution}: Pure-color anchor resolution ablation.
\item Sec.~\ref{app:color-shift}: Editing performance as a function of source-to-target color distance and on unnatural target colors.
\item Sec.~\ref{app:user-study}: Human preference study protocol and results.
\item Sec.~\ref{app:string-check}: Training and evaluation separation check.
\item Sec.~\ref{app:data-samples}: Additional training-data examples.
\end{list}

\section{Relation to Existing Color Benchmarks}
\label{app:benchmark-relation}

Existing benchmarks expose related weaknesses in color-controllable generation. ColorBind introduces CompColor for compositional color binding~\citep{shomerchai2025colorbind}, GenColorBench broadens evaluation to many color-control settings including numerical colors~\citep{butt2025gencolorbench}, and ColorConceptBench targets implicit color-concept understanding~\citep{ruan2026colorconceptbench}. ACBench complements these efforts by focusing on explicit 24-bit hex-to-object fidelity for both generation and editing.

\textbf{GenColorBench NCU.}
We also evaluate under the official GenColorBench Numerical Color Understanding (NCU) protocol~\citep{butt2025gencolorbench}: four images per RGB/HEX prompt at each model's default sampling steps and resolution, GroundingDINO+SAM2 localization, OneHue dominant-color extraction, and CIEDE2000 scoring with the released color-neighborhood tolerance. Table~\ref{tab:gencolorbench-ncu} reports the result alongside the per-category scores published by NumColor~\citep{butt2026numcolor}, whose evaluation setting we follow.

\begin{table}[!htbp]
\centering
\small
\setlength{\tabcolsep}{3pt}
\renewcommand{\arraystretch}{1.15}
\renewcommand{\PAtabwidth}{0.72\linewidth}
\begin{tabular*}{\PAtabwidth}{@{\extracolsep{\fill}}lrrrr@{}}
\toprule
Model & L1 & L3 & CSS3/X11 & Avg. $\uparrow$ \\
\midrule
\texttt{FLUX.2-4B} Base & 34.71 & 34.64 & 32.64 & 34.00 \\
\textbf{\ours{}} & \textbf{58.15} & \textbf{58.54} & \textbf{56.98} & \textbf{57.89} \\
FLUX + NumColor$^{\dagger}$ & 55.71 & 48.02 & 51.96 & 51.90 \\
SD3.5 + NumColor$^{\dagger}$ & 51.72 & 43.35 & 46.87 & 47.31 \\
SD3 + NumColor$^{\dagger}$ & 49.23 & 39.46 & 47.19 & 45.29 \\
PixArt-$\Sigma$ + NumColor$^{\dagger}$ & 48.56 & 40.83 & 43.90 & 44.43 \\
PixArt-$\alpha$ + NumColor$^{\dagger}$ & 44.93 & 36.17 & 38.51 & 39.87 \\
\bottomrule
\end{tabular*}

\caption{GenColorBench NCU, following the evaluation setting of NumColor~\citep{butt2026numcolor}. L1 and L3 denote ISCC-NBS levels, and Avg. is their arithmetic mean with CSS3/X11. $\dagger$: scores quoted from NumColor~\citep{butt2026numcolor}, Table~1, not rerun here.}
\label{tab:gencolorbench-ncu}
\end{table}

On the same FLUX.2-4B backbone, finetuning improves mean NCU from 34.00 to 57.89, a gain of 23.89 points, with improvements in all three categories. Our mean also exceeds the published FLUX + NumColor result (51.90) by 5.99 points. GenColorBench NCU remains a single-object T2I numerical-color test. ACBench also covers two-object binding and image editing, while CompColor provides an independent compositional-binding diagnostic.

\section{ACBench Construction and Evaluation}
\label{app:benchmark-prompts}

\textbf{Task composition and prompts.}
ACBench-T2I contains 1000 prompts over common object categories: 500 specify one object and one target hex color, and 500 specify two objects with distinct target hex colors. ACBench-Edit contains 500 real-image recoloring prompts, each specifying one target object and its requested hex color. The prompt forms are illustrated in Section~\ref{sec:experiments}. We check these prompts for exact-string overlap with Paint-500K in Appendix~\ref{app:string-check}.

\textbf{Editing images and masks.}
The 500 editing source images come from an internal dataset of high-quality real images that does not overlap with the Paint-500K source images. Masking follows Section~\ref{sec:experiments}, where we reuse one manually inspected source mask so that the evaluated region stays fixed across methods.

\textbf{Color targets and score interpretation.}
We sample target colors at random from the 24-bit RGB space and attach each target hex to its corresponding object, with distinct target colors in two-object prompts. Scoring uses the MAE-to-score mapping of Section~\ref{sec:experiments}, and we analyze its sensitivity to the color estimator and to localization in Appendix~\ref{app:acbench-reliability}.

\textbf{Relation to prior evaluation protocols.}
Following the object-centric evaluation approach of CompColor~\citep{shomerchai2025colorbind} and GenColorBench~\citep{butt2025gencolorbench}, we localize the requested object before assessing its color. ACBench uses its own SAM3 and sRGB-MAE protocol. For independent evaluation, we use the released CompColor evaluator without modification and the official GenColorBench NCU protocol, including its dominant-color extraction and CIEDE2000-based scoring (Appendices~\ref{app:compcolor-transfer} and~\ref{app:benchmark-relation}).

\subsection{Baseline Settings and Result Sources}
\label{app:baseline-settings}

For reproduced baselines, we follow the default configurations of their official open-source implementations and Hugging Face releases, except for the matched FLUX.2-4B editing CFG of 2.0 stated in the main paper.

\textbf{Prompt interfaces and result provenance.}
ACBench specifies numerical target colors, while CompColor uses named-color prompts unless a row is marked ``Hex.'' The latter replaces the color names with their canonical hex values while preserving object compositions. Color-specialized methods retain their respective model backbones and control interfaces. Among them, ColorWave is our reproduction, as marked in Table~\ref{tab:main-results}. We quote the SD1.5 and FLUX.1 CompColor scores in Table~\ref{tab:main-results} from ColorBind, as marked on those cells, but evaluate both models on ACBench ourselves. Additional quoted CompColor baselines appear in Appendix~\ref{app:compcolor-transfer}. NumColor NCU results are arithmetic means of its three published category scores, rather than ACBench reruns (Appendix~\ref{app:benchmark-relation}). Cross-model results compare complete systems, whereas same-backbone finetuning and ablation comparisons assess our training recipe.

\section{CompColor Hex Protocol}
\label{app:compcolor-transfer}

\textbf{Benchmark background.} CompColor is the compositional color-binding benchmark released by ColorBind~\citep{shomerchai2025colorbind}, originally designed to expose a failure mode of T2I models: when a prompt mentions \emph{multiple} colored objects, generators often leak color across objects, swap the two assignments, or collapse both objects to a shared dominant hue. CompColor isolates this failure mode by holding the textual structure fixed, varying only the (color, object) bindings, and measuring per-object color fidelity. Each prompt follows the template ``a \{color1\} colored \{object1\} and a \{color2\} colored \{object2\}''.

\textbf{Color palette and pair construction.} The released benchmark draws color words from a fixed \emph{named-color} palette (e.g., \colorterm{FF6347}{tomato}, \colorterm[white]{4169E1}{royal-blue}, \colorterm{E0FFFF}{light-cyan}, \colorterm{FF69B4}{hot-pink}) whose canonical RGB triplets are known. For each pair of color names, the perceptual distance between their canonical sRGB values is computed in CIELAB space (denoted $\Delta E_{ab}^{*}$ in the original paper). Color pairs are then partitioned by this distance into a \textit{Close} split, where the two colors are perceptually similar and easy to confuse (e.g., \colorterm{87CEEB}{sky-blue} vs.\ \colorterm{E0FFFF}{light-cyan}), and a \textit{Distant} split, where the two colors are perceptually well-separated (e.g., \colorterm{87CEEB}{sky-blue} vs.\ \colorterm{FF69B4}{hot-pink}). The benchmark also includes two \textit{Single} subsets (one color $\times$ one object), used to verify that single-object color rendering still works in isolation.

\textbf{Object compositions.} Object pairs are sampled from a curated list of common everyday categories (animals, vehicles, garments, household items, food) so that the requested colors are physically plausible and segmentable. The original benchmark fixes both color words and object words \emph{up front} and ships the prompts as a closed set, so all baselines are evaluated on exactly the same prompts and seeds. We respect this convention: our hex protocol modifies only the color tokens, never the object tokens or pair structure.

\textbf{Evaluation metric.} The original CompColor metric estimates, for each generated image, whether each requested object was rendered in the requested color. In practice this combines an object localizer (a segmenter applied to each object phrase) with a color-similarity check between the segmented region and the target RGB value. The released scores are normalized to a 0 to 1 scale, where higher means better object-level color binding. For CompColor results that we evaluate, including our named-color and hex-translated rows, we run the ColorBind open-source evaluation codebase~\citep{shomerchai2025colorbind} without modification. Concretely, the pipeline uses LangSAM, which combines Grounding DINO~\citep{liu2023grounding} with SAM~\citep{kirillov2023sam}, to locate each object from its text label and produce a segmentation mask, and then scores color with the original k-means-plus-$\Delta E_{\mathrm{CMC}}$ protocol.

\textbf{Hex translation.} We keep the original object compositions and the \textit{Single}/\textit{Close}/\textit{Distant} splits untouched, and replace only the color words in each prompt with explicit 24-bit hex strings derived from the canonical RGB value of the original color name (e.g., \colorterm{FF6347}{tomato} $\to$ \hexcolor{FF6347}, \colorterm[white]{4169E1}{royal-blue} $\to$ \hexcolor[white]{4169E1}). Named-color baselines and our hex-conditioned rows therefore describe the same perceptual targets, so within-model differences isolate the change from color words to hex specifications. This is a stricter test than CompColor's named-color setting: hex strings remove the explicit named-color cue (``\colorterm{FF6347}{tomato}'' suggests a reddish hue), while retaining any numerical-color knowledge already present in the pretrained text encoder.

\textbf{Subset aggregation.} The released benchmark provides two single-object subsets evaluated separately in the original paper. To match the score format used by the most recent baseline tables, we average them into a single \textit{Single} column in Table~\ref{tab:main-results} and report the \textit{Close} and \textit{Distant} columns as is. In Table~\ref{tab:compcolor-extra} we further report the unweighted mean of Single, Close, and Distant as an \textit{Avg} column, matching the original paper's averaging convention.

\textbf{Three-object extension.} CompColor was originally proposed as a two-object compositional benchmark, and the released baseline scores cover only this setting, which is the one reported in the main paper. We also evaluate the same wrapped-hex interface on CompColor's three-object extension (Table~\ref{tab:compcolor-3obj}).

\begin{table}[!htbp]
\centering
\small
\setlength{\tabcolsep}{5pt}
\renewcommand{\arraystretch}{1.12}
\resizebox{\ifdim\width>\linewidth\linewidth\else\width\fi}{!}{%
\begin{tabular}{lcc}
\toprule
Protocol & \ours{} & \texttt{FLUX.2-4B} Base \\
\midrule
Valid disjoint masks & 0.763 & 0.362 \\
Missing/overlapping masks score zero & 0.506 & 0.210 \\
\bottomrule
\end{tabular}

}
\caption{CompColor three-object extension evaluated under the wrapped-hex interface.}
\label{tab:compcolor-3obj}
\end{table}

\ours{} remains ahead under both protocols. The evaluator scores the designated target object, so this experiment measures color control in scenes containing three colored objects rather than strict three-way joint correctness.

\textbf{Score format.} All CompColor hex-protocol scores follow the original 0 to 1 convention, while ACBench scores follow our 0 to 100 percentage-point scale. The two benchmarks are not numerically comparable in Table~\ref{tab:main-results}: CompColor measures \emph{compositional binding} on a closed prompt set, while ACBench measures \emph{object-level hex color fidelity} on broader generation and editing protocols.

\textbf{Full original baseline roster.} The main paper Table~\ref{tab:main-results} keeps the modern T2I/edit baselines that overlap with ACBench so that the same backbone family can be tracked across both benchmarks. Table~\ref{tab:compcolor-extra} reports the remaining original CompColor baselines released by ColorBind~\citep{shomerchai2025colorbind}, namely Attend-and-Excite~\citep{chefer2023attendandexciteattentionbasedsemanticguidance}, Structured Diffusion~\citep{feng2022training}, SynGen~\citep{NEURIPS2023_0b08d733}, RichText~\citep{ge2023richtext} and Bounded Attention~\citep{dahary2024yourselfboundedattentionmultisubject}, with the same (Single, Close, Distant, Avg) format as the original paper. We copy these rows directly from the release, which uses named-color prompts only.

\begin{table}[!htbp]
\centering
\small
\setlength{\tabcolsep}{2.5pt}
\renewcommand{\arraystretch}{1.15}
\renewcommand{\PAtabwidth}{0.78\linewidth}
\begin{tabular*}{\PAtabwidth}{@{\extracolsep{\fill}}lrrrr@{}}
\toprule
Method & Single $\uparrow$ & Close $\uparrow$ & Distant $\uparrow$ & Avg $\uparrow$ \\
\midrule
SD 1.4                          & 0.51 & 0.38 & 0.28 & 0.39 \\
SD 2.1                          & 0.61 & 0.33 & 0.26 & 0.40 \\
Attend-and-Excite     & 0.55 & 0.46 & 0.38 & 0.46 \\
Structured Diffusion                                         & 0.58 & 0.39 & 0.39 & 0.45 \\
SynGen                                                   & 0.45 & 0.49 & 0.56 & 0.50 \\
RichText                                                         & 0.29 & 0.35 & 0.30 & 0.31 \\
Bounded-Attention              & 0.43 & 0.55 & 0.29 & 0.42 \\
\bottomrule
\end{tabular*}

\vspace{0.5em}
\caption{Remaining original CompColor baselines from ColorBind~\citep{shomerchai2025colorbind}, omitted from the main paper Table~\ref{tab:main-results}. Scores follow the released 0 to 1 convention.}
\label{tab:compcolor-extra}
\end{table}

\section{Training Data and Mixture}
\label{app:training-data}
Paint-500K contains 500K color-control samples built from an internal, non-public collection of high-quality real images. It is organized into two streams: 400K T2I generation samples and 100K editing samples. The T2I stream contains 100K single-object color generation samples and 300K multi-object color generation samples. Rows include target image paths, object prompts, source image paths for editing, object masks, target hex colors, and dominant-color metadata. We use Seed1.8~\citep{bytedanceseed2026seed18} as the VLM for grounding, filtering, and recaptioning, and use Qwen-Image-Edit-2511~\citep{wu2025qwenimage,qwen2025qwenimageedit2511} with the 4-step Qwen-Image-Edit-2511-Lightning LoRA~\citep{lightx2v2026qwenimageeditlightning} as the editing model.

Final training uses three streams in total: Paint-500K T2I generation, Paint-500K editing, and auxiliary pure-color grounding. Each training batch contains 30 real-image T2I samples, 12 pure-color samples, and 30 editing samples, giving a total batch size of 72. We set $\lambda_{\mathrm{t2i}}=\lambda_{\mathrm{edit}}=\lambda_{\mathrm{rgb}}=1$ in all experiments. The auxiliary pure-color stream contains 10K solid-color samples at $512\times512$ resolution, which we train only at timesteps $t\in[0.8,1.0]$ in the final recipe. All 4B finetuning and ablation rows use Adam, learning rate $2\times10^{-5}$, and 4000 steps. T2I generation with \texttt{FLUX.2-klein-base-4B} uses the default classifier-free guidance of 4.0 unless otherwise noted, and all FLUX.2-4B editing comparisons, including the base model and finetuned variants, use the same CFG of 2.0.

\textbf{Construction details.}
For instance grounding, each VLM proposal contains a text label, a short region description, and a bounding box. SAM3 then segments each grounded box so that we extract colors from object pixels. We normalize the CIELAB channels to $[0,1]$ and cluster masked pixels with MeanShift using bandwidth $0.05$. Two filters then decide which colors are retained: each retained cluster must cover at least $15\%$ of the object pixels (\texttt{COLOR\_MIN\_RATIO}${}=0.15$), and the retained clusters must jointly cover at least $90\%$ of the object pixels (\texttt{INSTANCE\_KEEP\_RATIO}${}=0.90$), or we discard the instance. Small highlight, shadow, or decoration regions are therefore removed, while multiple substantial colors on one object can be retained. When a single dominant color is required, we take the retained cluster with the largest pixel mass and convert it to a 24-bit sRGB hex label. In a spot check, we randomly inspected 100 Paint-500K samples, and 96 of their dominant-color labels agreed with the human-perceived dominant object color. In a second check, five annotators independently assessed target-object colors across the color gamut, and the MAE computed from their annotations differed from the MAE of the automatic extraction by 15.37\%.

When several objects share the same semantic label, VLM grounding returns separate bounding boxes for each instance, including same-category objects with different colors (Figure~\ref{fig:data-pipeline}), and we extract colors independently inside each box. During recaptioning, we feed the original caption, grounded layout, text labels, and extracted hex colors back to the VLM, which aggregates the localized instances, colors, and spatial relations into a complete caption and inserts exact \texttt{<color>\#HEX</color>} spans into the relevant noun phrases. These captions are the multi-object T2I prompts and can describe, for example, a red object on the left and a green object of the same category on the right. For single-object T2I samples, we start from instances whose dominant color covers a large fraction of the mask, randomly expand the instance box, crop a local image region, and recaption the crop around the selected object and its dominant hex color.

For editing data, we keep a synthesized source-target pair only when the mask-outside change is small enough to preserve the surrounding scene and the mask-inside change is large enough to verify a meaningful color edit. VLM verification and recaptioning further require that the target object remains identifiable, the requested color change is visually grounded, and any visible texture, material, pattern, or shape changes are reflected in the final instruction while preserving the exact target hex string.

\section{Pure-Color Generation Trajectories}
\label{app:timestep-vis}

\begin{figure}[ht]
\centering
\includegraphics[width=\linewidth]{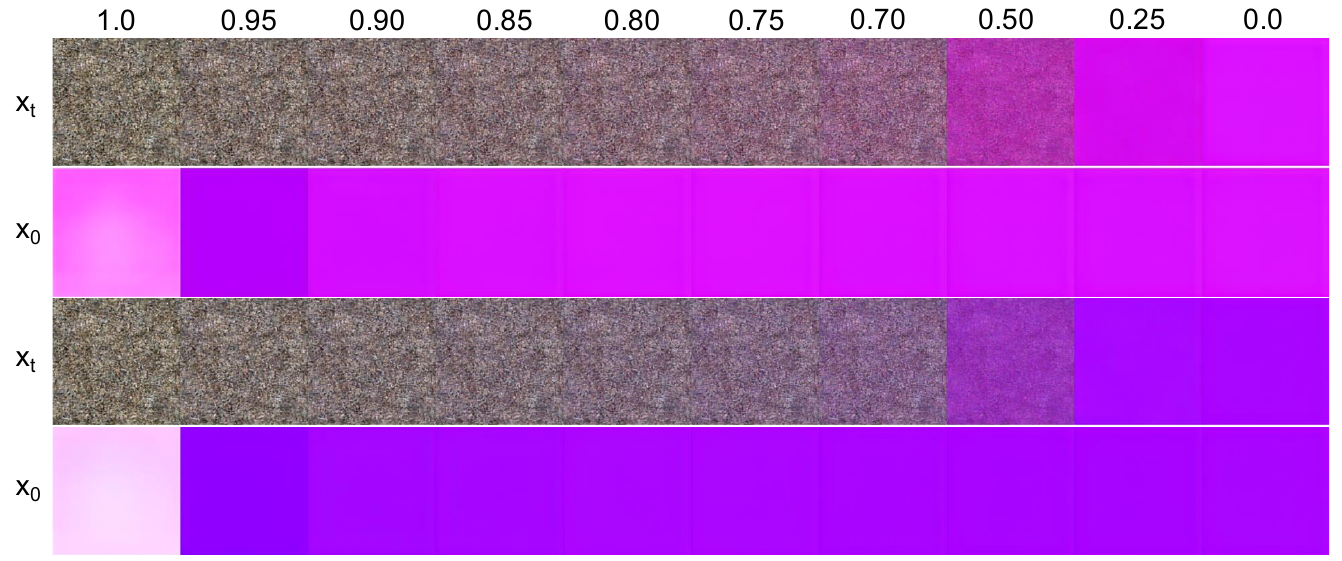}
\caption{\textbf{Early color stabilization in pure-color generation.} Two trajectories visualize noisy states (rows labeled $x_t$) and corresponding single-step denoised predictions (rows labeled $x_0$) at decreasing noise levels. In these examples, the predicted color is already visually stable near $t\approx 0.8$.}
\label{fig:timestep-vis}

\end{figure}

Figure~\ref{fig:timestep-vis} explores when color emerges in two pure-color generation trajectories, giving a qualitative motivation for high-noise anchor supervision. The downstream evidence comes from the gate ablation in Table~\ref{tab:gate-results}, where $t_{\mathrm{gate}}=0.8$ improves both generation and editing over ungated anchors.

\section{Prompt Templates and Data-Generation Tools}
\label{app:prompt-templates}

We used fixed prompt templates for VLM-based caption refinement, instance grounding, and edit-instruction refinement. The color-caption refinement system prompt was:

\begin{Verbatim}[fontsize=\scriptsize,breaklines=true,breakanywhere=true]
Role: You are a visual data refinement expert specializing in integrating
precise color attributes into image captions for VLM training.

Task: Integrate detected objects and their specific HEX color codes into a
global caption. Use the provided bounding box coordinates only to determine
spatial relationships and distinguish between multiple objects, but do not
include the numerical coordinates in the final output.

Constraints & Rules:

No Bbox in Output: Do not include any coordinate values (x1, y1, x2, y2)
or bbox IDs in the final caption. Use them only to infer positions
(e.g., "on the left," "in the background").
Preserve Context: Maintain the original global caption's narrative flow
and structure.
Disambiguation: If multiple similar objects exist, use their relative
spatial positioning to make the description unique
(e.g., "The <color>#HEX</color> bottle on the right").
Natural Integration: Insert the color tags directly before or after the
object they describe so the sentence remains grammatically fluent.
Return Format: "[Final enhanced caption text with <color>HEX code</color>
tags and no coordinates]"
Strict HEX Usage: You must use the EXACT HEX color codes provided in the
input. Do not invent or hallucinate colors that are not in the input
(e.g., #000000, #FFFFFF).

Example for your reference:
User Input:
Global Caption: A cat sitting on a sofa near a lamp.
Bounding Boxes:
id: 1 | label: cat | color: #4A4A4A | region_caption: a cat |
coords: [100, 200, 300, 400]
id: 2 | label: sofa | color: #F5F5DC | region_caption: a sofa |
coords: [0, 150, 800, 900]
id: 3 | label: lamp | color: #FFD700 | region_caption: a lamp |
coords: [700, 50, 850, 500]

Desired Output:
"A <color>#4A4A4A</color> cat is sitting on a <color>#F5F5DC</color>
sofa, positioned next to a <color>#FFD700</color> lamp on the right."
\end{Verbatim}

For instance grounding, we used the following prompt template:

\begin{Verbatim}[fontsize=\scriptsize,breaklines=true,breakanywhere=true]
Role: You are a visual instance grounding expert for precise color-control
data construction.

Task: Given an image and its global caption, identify caption-relevant
object instances that can support reliable object-level color labeling.
Return only objects that are visually present, localizable, and useful for
color extraction. For each instance, return a normalized bounding box.

Constraints & Rules:

Caption Faithfulness: Prefer objects explicitly mentioned in the caption.
You may include visually salient objects not mentioned in the caption only
when they are necessary to preserve the scene context.
Instance Separability: Split multiple similar objects into separate
instances when they can be distinguished by position, size, or appearance.
Spatial Disambiguation: Use relative positions such as "left", "right",
"front", "background", "upper", or "lower" to disambiguate instances.
No Hallucination: Do not output objects that are not clearly visible.
Color Suitability: Prefer objects with a coherent dominant surface color.
Avoid transparent, reflective, heavily shadowed, tiny, or highly textured
objects when their color cannot be reliably measured.
Normalized Bbox: Return each bounding box as normalized coordinates on a
0 to 1000 scale, in the order x1, y1, x2, y2.
Bbox Tags: The bbox value must be wrapped with <bbox></bbox>, for example
"bbox": "<bbox>[120, 85, 640, 730]</bbox>".
No Coordinates in Captions: Coordinate values must appear only in the
"bbox" field, not inside natural language captions or descriptions.
Return Format: Return only a valid JSON list. Each item must contain
"label", "instance_caption", "visual_description",
"spatial_description", and "bbox".
Example:
[
  {
    "label": "sky",
    "instance_caption": "the blue sky in the background",
    "visual_description": "a large coherent sky region",
    "spatial_description": "upper background",
    "bbox": "<bbox>[0, 0, 1000, 420]</bbox>"
  }
]
\end{Verbatim}

For edit-instruction refinement, we used the following prompt:

\begin{Verbatim}[fontsize=\scriptsize,breaklines=true,breakanywhere=true]
You are an image edit prompt enhancement assistant.

You are given:
- an original edit prompt,
- a source image,
- an edited image,
- and the target object label.

Your task is to enhance the original edit prompt, not rewrite it from scratch.

Rules:
1. Preserve the original editing intent.
2. Always preserve the original hex color value exactly as written in the
original prompt.
3. Do not replace the hex color with a natural language color name.
4. Improve wording so the prompt is natural, concise, and fully in English.
5. Keep the enhanced prompt under 30 words.
6. Mention texture, material, pattern, or shape changes only if they are
clearly visible in the edited image compared with the source image.
7. Do not add extra changes unless they are visually clear.
8. If the original prompt already matches the edit well, make only minimal
improvements.
9. If image quality is too poor to judge reliably, return exactly: error
10. Return only the enhanced prompt as one short sentence, or exactly: error
11. The output must contain the same hex color string from the original
prompt exactly once.
12. Do not output explanations.
\end{Verbatim}

\section{Evaluator Settings}
\label{app:evaluator-settings}

For both ACBench-T2I and ACBench-Edit, SAM3~\citep{carion2025sam3} localizes the target object, and we compute scores from the masked pixels: in T2I we segment the generated target object from the prompt, and in editing we segment the source object and reuse the source mask on the edited image without dilation. The predicted color is the mean RGB value inside the mask, and the final error is its channel-wise MAE to the target hex color. Samples with failed target localization receive zero score. Like CompColor, whose evaluator uses LangSAM and its released color-scoring protocol (Appendix~\ref{app:compcolor-transfer}), the evaluation is object-level, but the main paper metric replaces its thresholded LAB accuracy with the continuous hex-fidelity score defined in Section~\ref{sec:experiments}. Object-centric evaluation with pretrained detectors or segmenters has established precedents in GenEval~\citep{ghosh2023geneval}, T2I-CompBench~\citep{huang2023t2icompbench}, Grounded SAM~\citep{ren2024groundedsam}, and TokenCompose~\citep{wang2024tokencompose}.

\section{ACBench Reliability}
\label{app:acbench-reliability}

ACBench's primary score uses mean RGB inside the target mask followed by channel-wise MAE. Region averages can be affected by shading, material, and mask boundaries, so we test whether the paper's ranking depends on this particular estimator, on SAM3 thresholds, or on localization failures.

\textbf{Region-statistic robustness.}
We recompute seven T2I model/configuration comparisons and three editing models over 4{,}656 object instances with five estimators: Mean RGB (paper metric), MeanShift dominant RGB, Median RGB, Median Lab, and Dominant Lab (largest CIELAB $k$-means cluster). The alternatives cover the color-quantization and RGB/CIELAB evaluation used by ColorBind~\citep{shomerchai2025colorbind} and the dominant-hue Lab evaluation used by GenColorBench~\citep{butt2025gencolorbench}. Table~\ref{tab:region-stat-t2i} reports T2I scores on the 0 to 1 scale used by the robustness script, which is the paper's 0 to 100 convention divided by 100.

\begin{table}[!htbp]
\centering
\small
\setlength{\tabcolsep}{3pt}
\renewcommand{\arraystretch}{1.12}
\begin{tabular*}{\linewidth}{@{\extracolsep{\fill}}lccccc@{}}
\toprule
Model / Configuration & Mean RGB & MeanShift & Median RGB & Median Lab & Dominant Lab \\
\midrule
\ours{}, $t_{\mathrm{gate}}=0.8$ & 0.6858 & 0.6954 & 0.7341 & 0.7254 & 0.7228 \\
\ours{}, $t_{\mathrm{gate}}=0.9$ & 0.6754 & 0.6800 & 0.7122 & 0.7176 & 0.7228 \\
\ours{}, $t_{\mathrm{gate}}=0.0$ & 0.6383 & 0.6744 & 0.7053 & 0.6965 & 0.6931 \\
\texttt{FLUX.2-9B} Base & 0.3815 & 0.3734 & 0.4082 & 0.3930 & 0.4118 \\
\texttt{FLUX.2-4B} Base & 0.3702 & 0.3713 & 0.3886 & 0.3853 & 0.3943 \\
Z-Image-Turbo & 0.3356 & 0.3395 & 0.3703 & 0.3546 & 0.3568 \\
Qwen-Image & 0.2521 & 0.2860 & 0.2975 & 0.2945 & 0.3003 \\
\bottomrule
\end{tabular*}

\caption{ACBench-T2I under alternative region-color estimators (0 to 1). The ordering is preserved for MeanShift, Median RGB, and Median Lab. Dominant Lab ties the two strongest variants at the displayed precision, but every estimator preserves their advantage over the base models.}
\label{tab:region-stat-t2i}
\end{table}

The same ranking holds for ACBench-Edit, where \ours{}, Base and Qwen-Image-Edit score 0.7557, 0.5890 and 0.5440 under Mean RGB and every alternative estimator preserves this order (Spearman $\rho{=}1.0$). The T2I gap between \ours{} and Base remains large under every estimator: $+0.3156$ (Mean RGB), $+0.3241$ (MeanShift), $+0.3455$ (Median RGB), $+0.3401$ (Median Lab), and $+0.3284$ (Dominant Lab).

\textbf{SAM3 threshold sensitivity.}
With all other evaluator settings fixed, we sweep the SAM3 detection threshold from $0.1$ to $0.9$ and four coverage gates $(\text{min cluster}, \text{min total})\in\{(0.15,0.70),\allowbreak(0.10,0.60),\allowbreak(0.05,0.30),\allowbreak(0,0)\}$, yielding $9\times4=36$ combinations. Table~\ref{tab:sam3-threshold} reports the default-coverage slice. \ours{} ranks above \texttt{FLUX.2-4B} Base at every threshold, with a margin from $+0.2482$ to $+0.3156$. Across the full 36-setting grid the gap ranges from $+0.2482$ to $+0.3308$.

\begin{table}[!htbp]
\centering
\small
\setlength{\tabcolsep}{5pt}
\renewcommand{\arraystretch}{1.12}
\resizebox{\ifdim\width>\linewidth\linewidth\else\width\fi}{!}{%
\begin{tabular}{cccc}
\toprule
SAM3 detection threshold & \ours{} & \texttt{FLUX.2-4B} Base & Difference \\
\midrule
$0.1$ to $0.5$ & 0.6858 & 0.3702 & $+0.3156$ \\
$0.6$ & 0.6837 & 0.3696 & $+0.3141$ \\
$0.7$ & 0.6760 & 0.3642 & $+0.3118$ \\
$0.8$ & 0.6623 & 0.3526 & $+0.3097$ \\
$0.9$ & 0.5659 & 0.3177 & $+0.2482$ \\
\bottomrule
\end{tabular}

}
\caption{ACBench-T2I under SAM3 detection-threshold sweep at default coverage $15\%/70\%$ (0 to 1).}
\label{tab:sam3-threshold}
\end{table}

\textbf{Localization-failure audit.}
To measure how often the target object cannot be localized, we set a SAM3 confidence threshold and score failures as zero. After verifying that high-confidence masks correctly cover the target on 20 random samples per model, we draw 600 images per model from the 1{,}000 ACBench-T2I outputs and measure failure rates (Table~\ref{tab:localization-failures}).

\begin{table}[!htbp]
\centering
\small
\setlength{\tabcolsep}{5pt}
\renewcommand{\arraystretch}{1.12}
\begin{tabular}{lcc}
\toprule
Model & Failures / 600 & Failure rate \\
\midrule
\texttt{FLUX.2-4B} Base & 25 & 4.17\% \\
\texttt{FLUX.2-9B} Base & 14 & 2.33\% \\
Qwen-Image & 10 & 1.67\% \\
Z-Image-Turbo & 1 & 0.17\% \\
\ours{} & 33 & 5.50\% \\
\bottomrule
\end{tabular}

\caption{Localization failure rates over 600 ACBench-T2I images per model.}
\label{tab:localization-failures}
\end{table}

Manual inspection attributes only $11\%$ of failures to SAM3, and most arise from the generator not producing a localizable target object. \ours{} has the highest failure rate among the listed models, so the zero score for failures does not favor it in this audit. Excluding all failures raises \ours{} from 68.58 to 70.93 without changing the model ranking. For ACBench-Edit, all models share the same human-verified source mask, so inter-model mask quality is matched by construction.

\section{Pure-Color Anchor Resolution}
\label{app:anchor-resolution}

Holding the remaining training recipe fixed and training for 4{,}000 steps, we vary only the solid-color anchor resolution (Table~\ref{tab:anchor-resolution}).

\begin{table}[!htbp]
\centering
\small
\setlength{\tabcolsep}{6pt}
\renewcommand{\arraystretch}{1.12}
\begin{tabular}{lcc}
\toprule
Anchor resolution & ACBench-T2I & ACBench-Edit \\
\midrule
$256\times256$ & 63.10 & 73.82 \\
$512\times512$ (default) & 68.58 & 75.57 \\
$1024\times1024$ & 69.52 & 76.03 \\
\bottomrule
\end{tabular}

\caption{Pure-color anchor resolution, with the rest of the recipe held fixed.}
\label{tab:anchor-resolution}
\end{table}

Dropping to $256\times256$ costs 5.48 points on ACBench-T2I and 1.75 on ACBench-Edit. Raising the resolution to $1024\times1024$ adds only 0.94 and 0.46 points at a higher training cost, so we keep $512\times512$ as the default.

\section{Editing under Large Color Shifts and Unnatural Colors}
\label{app:color-shift}

To examine editing under different source-to-target color shifts, we analyze a subset of ACBench-Edit. We divide this subset into four equal-sized quartiles by the CIEDE2000 distance between the estimated source-object color and the requested target, and separately isolate near-complementary cases where both colors are chromatic and the hue rotation exceeds $120^{\circ}$. Table~\ref{tab:color-shift} reports scores (0 to 100) on this subset, and Table~\ref{tab:main-results} reports the full-benchmark scores.

\begin{table}[!htbp]
\centering
\small
\setlength{\tabcolsep}{4.5pt}
\renewcommand{\arraystretch}{1.12}
\resizebox{\ifdim\width>\linewidth\linewidth\else\width\fi}{!}{%
\begin{tabular}{lccccc}
\toprule
Model & Q1 (nearest) & Q2 & Q3 & Q4 (farthest) & Near-comp. \\
\midrule
\ours{} & 78.8 & 77.1 & 79.2 & 88.9 & 86.9 \\
\texttt{FLUX.2-4B} & 67.6 & 64.2 & 62.9 & 56.3 & 50.8 \\
\bottomrule
\end{tabular}

}
\caption{ACBench-Edit scores (0 to 100) by source-to-target CIEDE2000 distance.}
\label{tab:color-shift}
\end{table}

\ours{} remains effective in every bucket and is strongest on the farthest quartile and the near-complementary subset, while the base model degrades as the required color shift grows. CompColor Distant under the hex-translated protocol shows the same pattern for target-pair contrast (Section~\ref{sec:experiments}).

\textbf{Unnatural target colors.} Among the 500 ACBench-Edit instructions, we identified 33 that recolor an object to an unnatural color. \ours{} scores 72.98 on this subset, close to its overall ACBench-Edit score of 75.57. The subset covers object recoloring only and contains no large scene regions such as a green sky.

\section{User Study}
\label{app:user-study}

We collect color-preference judgments from 15 participants on 80 randomly sampled ACBench-T2I prompts. Each trial shows a target swatch and two anonymized images from \ours{} and \texttt{FLUX.2-4B}. Across $15\times80=1200$ judgments, \ours{} is preferred in 55\%, tied in 32\%, and loses in 13\%, a non-tie win rate of $55/(55+13)=80.9\%$. Across 200 paired image-quality judgments, \ours{} wins 22\%, ties 58\%, and loses 20\%, giving a 52.4\% non-tie win rate. The color preferences agree in direction with the ACBench comparison.

\section{Training and Evaluation Separation}
\label{app:string-check}

We construct evaluation prompts independently, in formats different from the training captions. For the exact overlap check, we normalize the benchmark prompt strings and the object prompt, T2I prompt, and edit instruction fields of Paint-500K in the same way, by trimming whitespace, lowercasing, and collapsing internal whitespace. Under this normalization, no benchmark prompt exactly matches a training prompt.

\clearpage
\onecolumn
\section{Additional Training-Data Examples}
\label{app:data-samples}
We show 20 T2I examples and 20 editing pairs selected for visual diversity from the available training-data sample pools. For display, we translate the text from the original Chinese annotations and shorten the editing captions for readability, keeping the target hex values unchanged.

\begin{center}
\includegraphics[page=1,width=\linewidth,height=.80\textheight,keepaspectratio]{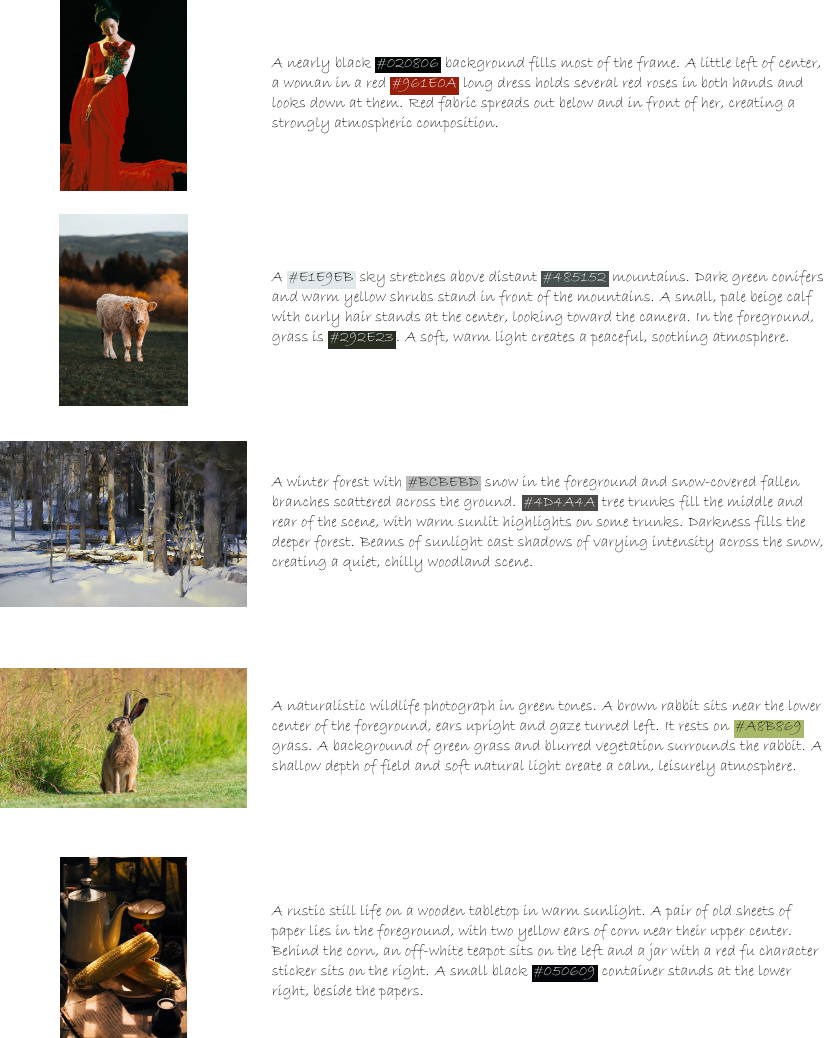}
\captionof{figure}{\textbf{T2I training-data examples (1/4).} Five image--caption examples with target hex colors.}
\label{fig:data-t2i-1}
\end{center}
\clearpage
\begin{center}
\includegraphics[page=2,width=\linewidth,height=.91\textheight,keepaspectratio]{figures/data_samples_t2i.pdf}
\captionof{figure}{\textbf{T2I training-data examples (2/4).} Five image--caption examples with target hex colors.}
\label{fig:data-t2i-2}
\end{center}
\clearpage
\begin{center}
\includegraphics[page=3,width=\linewidth,height=.91\textheight,keepaspectratio]{figures/data_samples_t2i.pdf}
\captionof{figure}{\textbf{T2I training-data examples (3/4).} Five image--caption examples with target hex colors.}
\label{fig:data-t2i-3}
\end{center}
\clearpage
\begin{center}
\includegraphics[page=4,width=\linewidth,height=.91\textheight,keepaspectratio]{figures/data_samples_t2i.pdf}
\captionof{figure}{\textbf{T2I training-data examples (4/4).} Five image--caption examples with target hex colors.}
\label{fig:data-t2i-4}
\end{center}
\clearpage
\begin{center}
\includegraphics[page=1,width=\linewidth,height=.91\textheight,keepaspectratio]{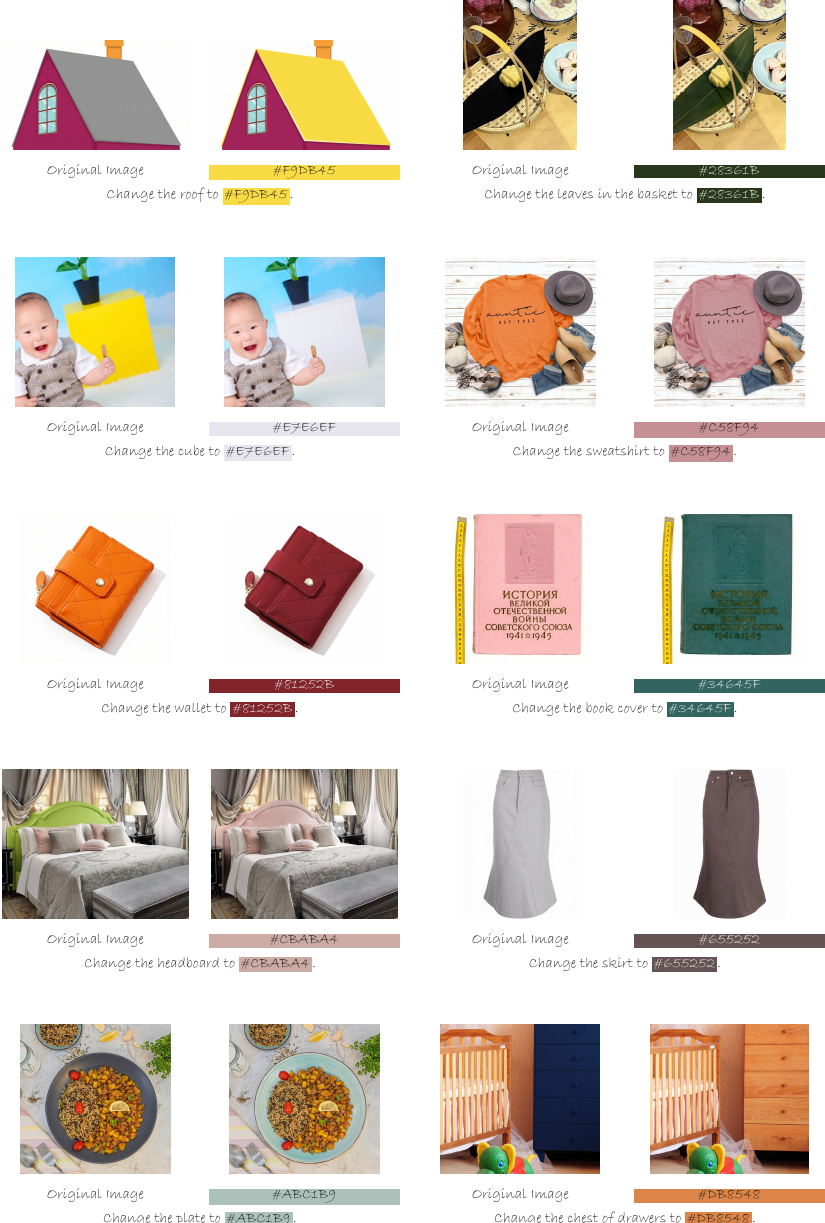}
\captionof{figure}{\textbf{Editing training-data examples (1/2).} Ten source--target pairs with short instructions. Each pair places the source on the left and the target on the right.}
\label{fig:data-edit-1}
\end{center}
\clearpage
\begin{center}
\includegraphics[page=2,width=\linewidth,height=.91\textheight,keepaspectratio]{figures/data_samples_edit.pdf}
\captionof{figure}{\textbf{Editing training-data examples (2/2).} Ten source--target pairs with short instructions. Each pair places the source on the left and the target on the right.}
\label{fig:data-edit-2}
\end{center}

\end{document}